\documentclass[10pt,a4paper]{article}

\usepackage{mathptmx}       
\usepackage{helvet}         
\usepackage{courier}        
\usepackage{type1cm}        
\usepackage{makeidx}         
\usepackage{graphicx}        
\usepackage{multicol}        
\usepackage[bottom]{footmisc}
\usepackage{url}
\usepackage{amsmath}
\usepackage{amssymb}
\usepackage{bm}
\usepackage{enumitem}

\usepackage[affil-it]{authblk}

\usepackage{xspace}
\usepackage{color}

\newlength{\smallimage}
        \definecolor{rel}{rgb}{.1,.6,.2}
        \definecolor{nrl}{rgb}{1,1,1}
        \definecolor{qim}{rgb}{1,1,1}

\newcommand{\rk}[1]{\bf{\color{red}#1}}

\newcommand{\bk}[1]{{\color{blue}#1}}

\makeatletter

\DeclareRobustCommand\onedot{\futurelet\@let@token\@onedot}
\def\@onedot{\ifx\@let@token.\else.\null\fi\xspace}

\def\eg{\emph{e.g}\onedot} 
\def\ie{\emph{i.e}\onedot}

\def\etal{\emph{et al}\onedot}

\newcommand{\mysec}[1]{section~\ref{#1}}
\newcommand{\fig}[1]{Figure~\ref{#1}}

\def\eq#1{Eq.~\ref{#1}}
\def\be{\begin{equation}}
\def\ee{\end{equation}}
\def\bea{\begin{eqnarray}}
\def\eea{\end{eqnarray}}
\def\ben{\begin{eqnarray*}}
\def\een{\end{eqnarray*}}

\def\bi{\begin{itemize}}
\def\ei{\end{itemize}}

\newcommand{\bt}[1]{\begin{tabular}{#1}}
\newcommand{\et}{\end{tabular}}

\def\g{\gamma}

\def\g{\gamma}
\def\S{\Sigma}
\def\s{\sigma}

\def\<{\langle}
\def\>{\rangle}

\makeindex             

\title{Document image classification, with a specific view on applications of patent 
images\footnote{To appear in M. Lupu \etal (eds.), Current Challenges in Patent 
Information Retrieval, second edition, 2016. Paper submitted in 2014.}}
\author{Gabriela Csurka}
\affil{Xerox Research Centre Europe, 6 chemin de Maupertuis \\
38240,  Meylan France \\
\texttt{Firstname.Lastname@xrce.xerox.com}
}

\begin{document}
\maketitle

\begin{abstract}
\, The main focus of this paper is document image classification  and retrieval, where  
we analyze and compare different parameters for the RunLeght Histogram (RL) and 
Fisher Vector (FV) based image representations. We do an exhaustive
experimental study  using different document image datasets, 
including the MARG benchmarks, two datasets built on customer data  and the images 
from the  Patent Image Classification task of the Clef-IP 2011. 
The aim of the study is to give  guidelines on
how to  best choose the parameters such that the same features 
perform well on different tasks. As an example of such need,  
we describe the Image-based Patent Retrieval task's
of  Clef-IP 2011, where we used the same image representation to predict the image type and
retrieve relevant patents.
\end{abstract}

\section{Introduction}
\label{sec:intro}

 {\em ``Before a patent can be granted, patent offices
perform thorough searches to ensure that no previous similar
disclosures were made. In the intellectual property 
terminology, such kind of searches are called prior art searches [...] 
Often, patent applications contain images that clarify 
details about the invention they
describe. Images in patents may be drawn by hand, by computer, or both, may
contain text, and are generally black-and-white 
(\ie not even monochrome). Depending on the technological area of a patent, 
images may be technical drawings
of a mechanical component, or an electric component, 
flowcharts if the patent
describes, for example, a workflow, chemical structures, 
tables, etc. When a patent expert browses through a list of search results given by a search engine,
he or she can very quickly dismiss irrelevant patents to the patent application by
just glancing at the images in the retrieved patents. The number of documents to
be looked at in more detail is thus greatly reduced.''} 
\cite{Piroi11}.

From this citation we can see that images are essential components of  a patent
as they illustrate key aspects of the invention.  However,  not every image in a patent 
has the same importance.  Indeed, for patents related to chemistry or to 
pharmaceutic inventions images containing 
chemical structures or gene sequences are the most important,
while  searching for similar drawings containing electronic circuits  
can help patent experts in physics and electricity. If a patent expert is 
looking for prior art given a query patent and the system 
retrieves patent based on visual similarities between all images of the query patent and 
in the patent database, the system might return non relevant patents based on
visual similarity between flowcharts or tables. This would not 
 necessarily help the prior art search process. On contrary, if only images of a 
 certain type are considered, the retrieval can be significantly improved as 
 shown in \cite{clefip11} where the retrieval accuracy when searching for the 
most similar drawings between patent images was much higher than the 
accuracy obtained when considering similarities between all images. However this requires first 
 to identify the image type  (such as drawing, flowchart, ...)  to be considered.

 In general manual annotation of the patent images according to their
type is  either non-existent or poor with many errors, therefore there
is a clear need to be able to predict the image type automatically. Hence,
one of the main focus of this paper is to consider 
patent image classification  according to image types as the ones identified and used
in the Patent Image Classification task of Clef-IP 2011~\cite{Piroi11}, namely 
{\em abstract drawing, graph,  flowchart, gene sequence, program listing, symbol, 
chemical structure, table and mathematics}. 
On the other hand, as similar image search (retrieval)  is another important aspect of 
patent based applications such as prior art search, 
 in the paper we also address image similarities and image based retrieval.  For both tasks, we consider mainly two popular 
image representations, the Fisher Vector~\cite{Perronnin07,PSM10}  
and the RunLenghts Histograms~\cite{ChanChang01,KSB2007,Gordo13,GPV13}, 
and  compare different parameter configurations for them in order to come 
with useful guidelines related to their choice independently of the targeted problem.

As patent  images can be seen as particular document images, instead of limiting
our study to patent images, we will  address  the problem in a  
more generic way, by questioning  what is a good 
representation in general for document images.  First, in  \mysec{sec:imprep}  
we   briefly revise  the most popular  document image representations. 
Then, after describing in 
\mysec{sec:dataset} the datasets considered for the study and the experimental setup,  
\mysec{sec:rlexp} and \mysec{sec:fvexp} will be devoted to an exhaustive 
 parameter comparison for RunLength and Fisher Vector 
image representations respectively and in  \mysec{sec:comb_rl_fv}  we discuss different combinations
of RL and FV. In \mysec{sec:challenge} we we describe the Image-based Patent Retrieval task's
of  Clef-IP 2011, where we used the same image representation to predict the image type and
retrieve relevant patents. Finally, we conclude the paper in \mysec{sec:conclusion}.

\section{Document Image Representations}
\label{sec:imprep}
In the last few years  different image representations were proposed to 
 deal with document image classification and retrieval 
 that do not rely on OCR \ie seeing a document page as an image. 
To mention a few, see for more examples \cite{ChenBlostein07,KSB2007,Gordo13}, 
Cullen \etal \cite{CHH1997} propose feature sets including  
densities of interest points, histogram of the size and density of connected components, 
vertical projection histograms etc.  In \cite{HDRT1998} a multi-scale density 
decomposition of the  page is used to produce fixed-length descriptors 
 constructed efficiently from integral images.  The  features vectors proposed 
in \cite{SD2001} are based on  text versus non-text percentage, column structure, 
content  area and connected components densities. Bagdanov and Worring 
\cite{BW2003} propose a representation based on density changes obtained with different
 morphological operations. In \cite{S2006} document images are described 
as a list of salient Viola-Jones based features. However, these features 
contain relatively limited amount of information and while they might perform well
on a specific dataset and task for which they were designed, they are not generic enough to
be able to handle various document class types, datasets and tasks. As early 
natural image representations, such as color histograms, 
were significantly outperformed and replaced 
by the  successful introduction of the bag of visual words (BOV) image
representation \cite{Sivic03,Csurka2004}, the RunLength  Histograms 
have shown to be more generic and hence better suited for document image 
representation\footnote{Note that since when the paper was written, with the recent success of the 
deep convolutional neural networks (CNNs), new, richer representations 
were proposed for natural images and applied also to document images \cite{lekang14,harleyetal15}. 
The comparison of those representations with FV and RL is subject of future work.}~\cite{ChanChang01,KSB2007,Gordo13,GPV13}.

In this work,  therefore we focus on one hand on 
the  RunLength  Histograms, on the other hand we  consider as alternative 
the Fisher Vectors~\cite{Perronnin07,PSM10} which is 
the most successful extension of BOV image representation. 
In the following sections we briefly describe how these features are extracted from a document
image and which are their main parameters that have to be considered in particular 
when we build the corresponding image signatures.

\begin{figure}[ttt]
\centering
\hspace{-0.5cm}\includegraphics[height=2.7cm]{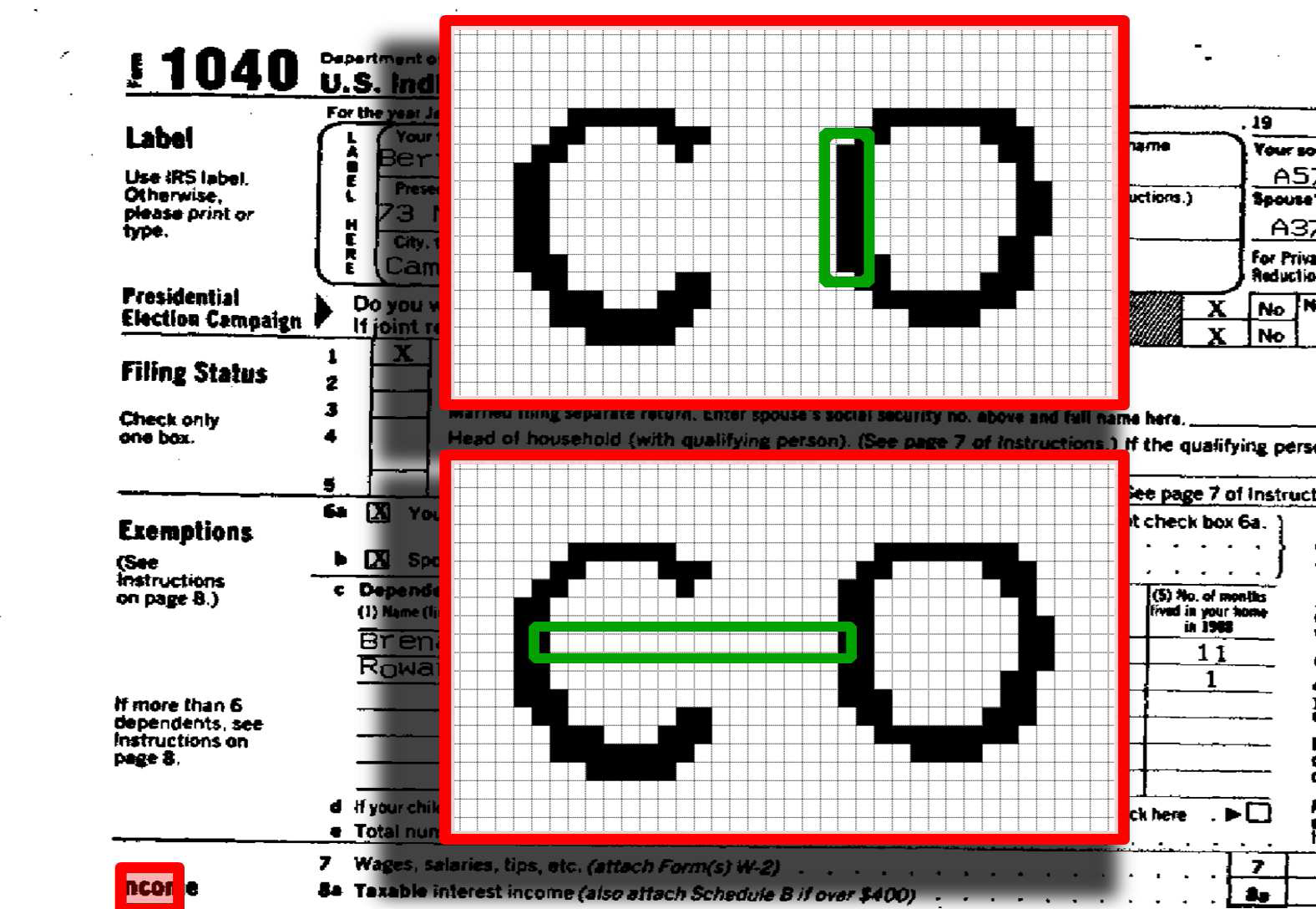}\hspace{0.5cm}
\includegraphics[height=2.7cm]{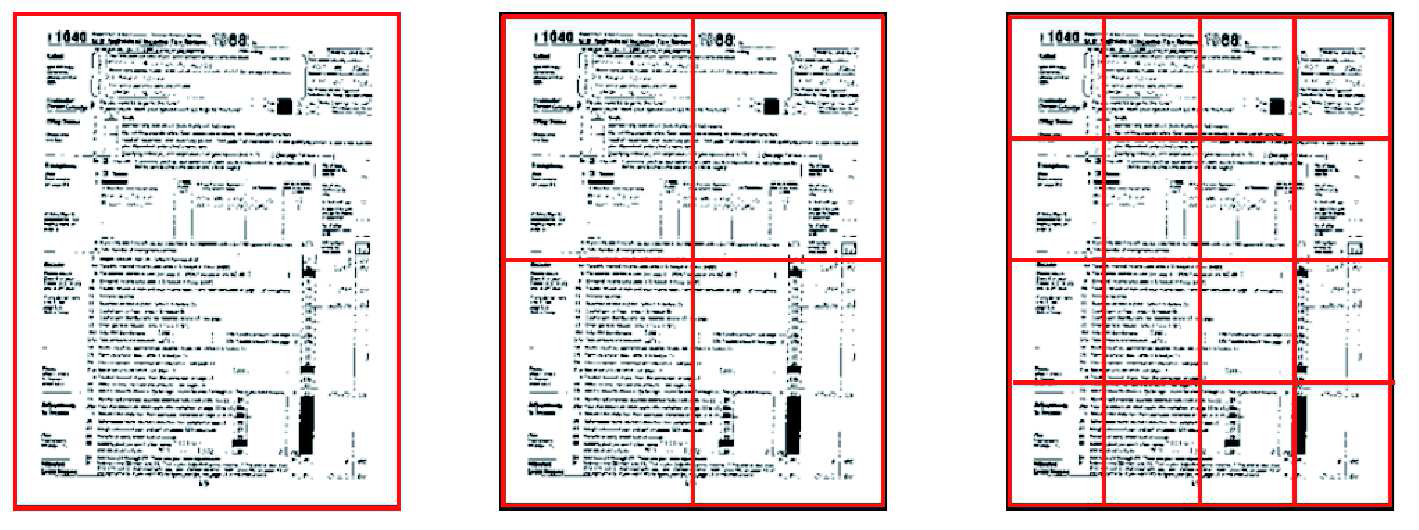}
\caption{Left: Examples of pixel runs. A vertical black run of length 7 (top) and
an horizontal white run of length 16 (bottom). 
Detail from a small region on the bottom-left corner. 
Right: A tree layer spatial pyramid. (Image courtesy of A. Gordo, from ~\cite{Gordo13}).}
\label{fig:RLex}
\end{figure}

\subsection{RunLength Histogram based Document Image Representation}
\label{sec:RL} 

The main idea of the RunLength (RL)  features is to encode sequences of 
pixels having the same value and going in the same direction (\eg vertical, horizontal 
or diagonal). The "run-length" is the length of those sequences (see as examples
the green  rectangles  in the Fig.\ref{fig:RLex}).  While we can consider sequences of similar 
gray-scale or even color values, considering only two levels has been proved to be sufficient 
to characterize  document images~\cite{Gordo13,GPV13}. Therefore, when needed, we first binarize 
the document images and we consider only runs of black and white pixels. In case of color
images, the  luminance channel is binarized. 

To do the binarization, we do a simple thresholding at 0.5 (where
 image pixels intensities are represented between 0 and 1). More complex binarization techniques 
 exist (see for example methods that participated in the DIBCO~\cite{Pratikakis12} and 
 HDIBCO~\cite{Pratikakis12} contests), however  testing the effect of different 
 binarization techniques  is out of the scope of this paper.  

On the binarized images, the number of  black pixel and white pixel runs are
collected into histograms.  To build these histograms, with the aim of being less sensitive to noises 
and small variations, we consider logarithmic  quantization of the lengths as
suggested in \cite{Gordo13,GPV13}:
\begin{equation*}
 [1], [2], [3-4], [5-8], [9-16] ,  \ldots, [\geq (2^q+1)].
\end{equation*}
Dealing with binary  images, this yields 2 histograms of length $Q=q+2$ per direction, 
one for the white pixels and one for the black pixels. We compute these runs in 4 directions, 
horizontal, vertical, diagonal and anti-diagonal,  
and concatenate the obtained histograms. An image (or image region) is then represented by this  
$4\times2\times Q$ dimensional feature called RunLength (RL) histogram.

These histograms can be computed  either on the whole image or on  image regions. 
In order to better capture information  about the page layout
we use a spatial pyramid~\cite{LSP06} with several layers such that  at each level the image is 
divided into $n \times n$ regions and the histograms  computed on these regions are concatenated. 
For example, in the case of a 3 layer pyramid  $1 \times 1,2\times 2,4\times 4$ illustrated 
in  Fig.\ref{fig:RLex}(right), we concatenate in total the RLs of 21 regions to obtain 
the final image signature. 

Finally, to be independent from the image size (number of pixel in the image) 
we  L1 normalize the signature followed by a component-wise power normalization\footnote{The 
component-wise power  normalization~\cite{PSM10} of a vector is such that each element $z$
is replaced by  $\mbox{sign}(z) |z|^{\alpha}$.} with $\alpha=0.5$
as in \cite{GPV13}. Note that a vector with positive elements having L1 norm equal to 1, after 
power normalization will have  L2 norm equal to 1.

\subsection{The  Fisher Vector  based Image representation}
\label{sec:bovFV}

\begin{figure}[ttt]
\centering
\includegraphics[height=3.4cm]{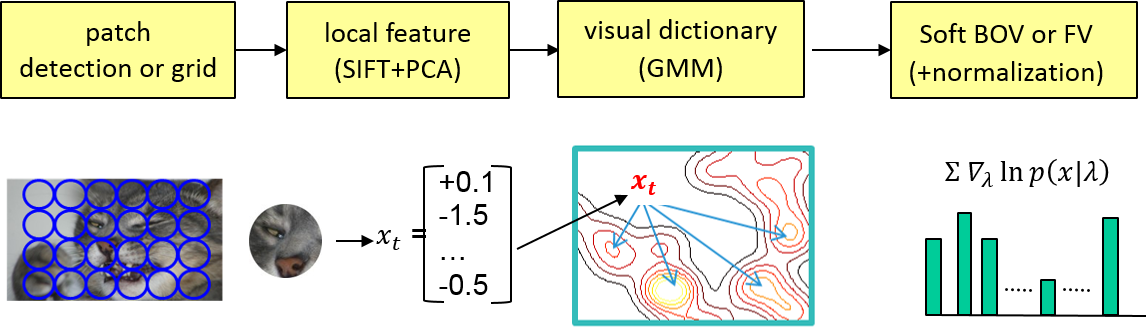}\hspace{0.5cm}
\caption{Illustration of the FV image representation pipeline.}
\label{fig:FVschema}
\end{figure}

The Fisher Vector \cite{Perronnin07} extends the bag-of-visual words (BOV) 
image representation by going beyond simple
counting (0-order statistics) as they encode higher order statistics about the distribution of local
descriptors assigned to visual words (see also Fig.\ref{fig:FVschema} illustrating the pipeline).
Similarly to the BOV, the FV depends on an 
intermediate representation: the visual vocabulary \cite{Sivic03,Csurka2004}. 
The visual vocabulary can be seen as a  probability density function (pdf)  which models 
the emission of  the low-level descriptors in the image. In our case  
we consider the Gaussian mixture model (GMM) to represent this density.

The  Fisher Vector characterizes  the set of low-level features (in our case SIFT features~\cite{lowe04ijcv}),
 $X_I=\{\bm x_t\}_{t=1}^{T}$ extracted from an image $I$ by deriving in which direction the 
parameters of the GMM model should be modified to best fit this particular
feature set. Assuming  independence, this  can be written as: 
\begin{equation}
G_{\lambda}(I)  = \frac{1}{T} \sum_{t=1}^{T} \nabla_{\lambda}
\log \left\{\sum_{n=1}^N w_n {\mathcal N}(\bm x_t | \mu_n,\Sigma_n)\right\}
\end{equation}
where $w_n$, $\mu_n$ and $\S_n$ denote
respectively the weight, mean vector and covariance matrix of the Gaussian $n$
and $N$ is the number of Gaussians in the mixture. To compare 
two images $I$ and $J$, a natural kernel on these gradients is  the Fisher Kernel
$K(I,J) = {G_{\lambda}(I)}^{\top} F_{\lambda}^{-1} G_{\lambda}(J)$, 
where $F_{\lambda}$ is the  Fisher Information Matrix. As $F_{\lambda}^{-1}$  
is symmetric and positive definite, it has a Cholesky decomposition
$L_{\lambda}^{\top} L_{\lambda}$ and  $K(I,J)$ can be rewritten as a 
dot-product between normalized vectors $\Gamma_{\lambda}$ where: 
\be
\Gamma_{\lambda}(I) =  L_{\lambda} G_{\lambda}(I)
\ee 
to which we refer as the {\em Fisher Vector} (FV) of
the  image $I$. 

Following \cite{Perronnin07,PSM10} where the covariance matrices in the GMM are 
assumed to be diagonal and using a diagonal closed-form approximation of  $F_{\lambda}$, 
we have:
\begin{eqnarray}
\Gamma_{\mu^d_n}(I)  & = & \frac{1}{T\sqrt{w_n}} \sum_{t=1}^T 
\g_n(\bm x_t) \left( \frac{\bm x^d_{t}-\mu^d_{n}}{\s^d_{n}} \right) \label{eqn:dm} , \\
\Gamma_{\s^d_n}(I) &  = & \frac{1}{T\sqrt{2 w_n}} \sum_{t=1}^T 
\g_n(\bm x_t) \left[ \frac{(\bm x^d_t-\mu^d_n)^2}{(\s^d_n)^2} -1 \right] 
\label{eqn:ds} 
\end{eqnarray}
where $\g_n(\bm x_t)=\frac{w_n {\mathcal N}(\bm x_t | \mu_n,\Sigma_n)}
{\sum_{j=1}^N w_j {\mathcal N}(\bm x_t | \mu_j,\Sigma_j)}$ and
$\s^d_{n}$ are the elements of the  diagonal $\Sigma_n$.
The final gradient vector $\Gamma_{\lambda}(I)$ is the concatenation of all
$\Gamma_{\mu^d_n}(I)$ and $\Gamma_{\s^d_n}(I)$,  where we 
ignore the gradients with respect to the weights. This vector is hence
$2ND$-dimensional, where D is the dimension of the low level features $\bm x_t$.

As  proposed in \cite{PSM10} we further apply  on $\Gamma_{\lambda}(I)$ a 
component-wise power  normalization a  component-wise power normalization~\cite{PSM10},
 followed by L2 normalization. 
Finally, similarly to the RunLengh features, to better 
take into account the document layout,  
we also consider similar spatial pyramids~\cite{LSP06} as in the case of RLs, \ie 
 dividing the image into several regions
at multiple layers and concatenating the region FVs.

\section{Datasets and the experimental setup}
\label{sec:dataset}

\begin{figure}[ttt]
\centering
\includegraphics[height=0.25\textwidth]{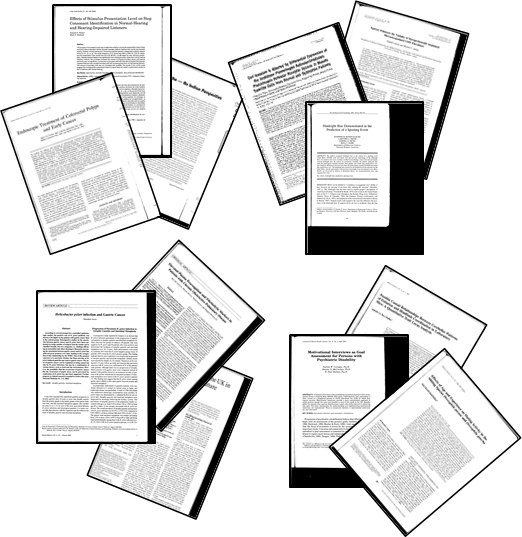} \;\;\;
\includegraphics[height=0.25\textwidth]{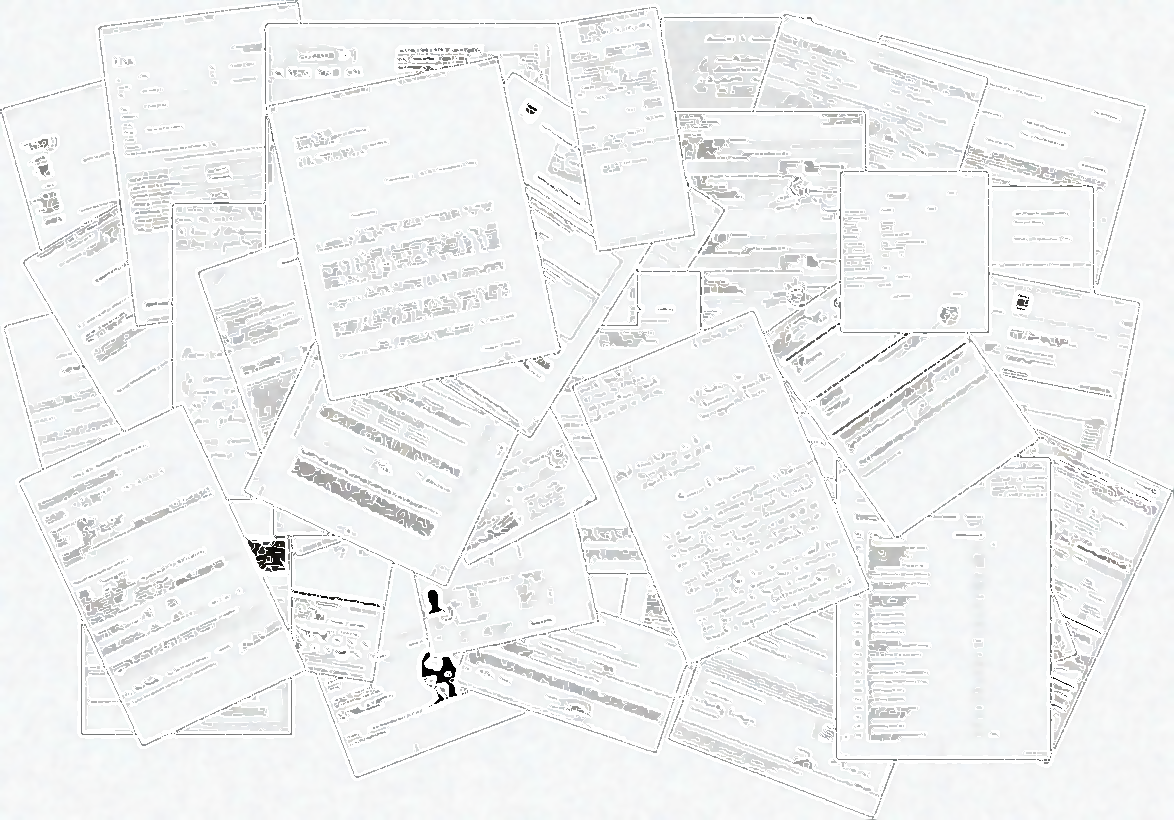} \;\;\;
\includegraphics[height=0.25\textwidth]{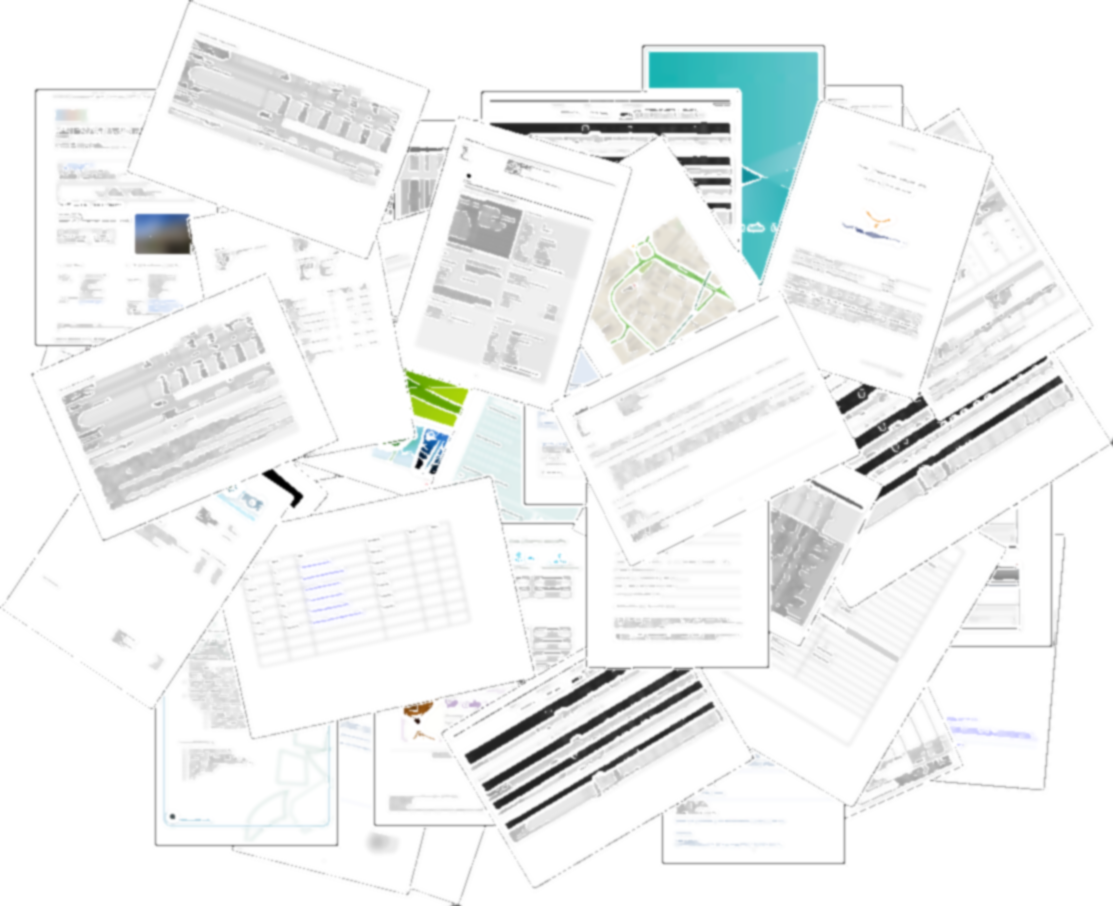} 
\caption{Example images  from four MARG class (left), and from the customer datasets IH1 (middle)  
and NIT (right). The images from customer datasets were intentionally blurred to 
keep the actual content of the documents confidential. Nevertheless, we can see 
the visual variability of the documents within these datasets. }
\label{fig.BYTELEX}
\end{figure}

\begin{figure}[ttt]
\centering
\includegraphics[width=0.9\textwidth]{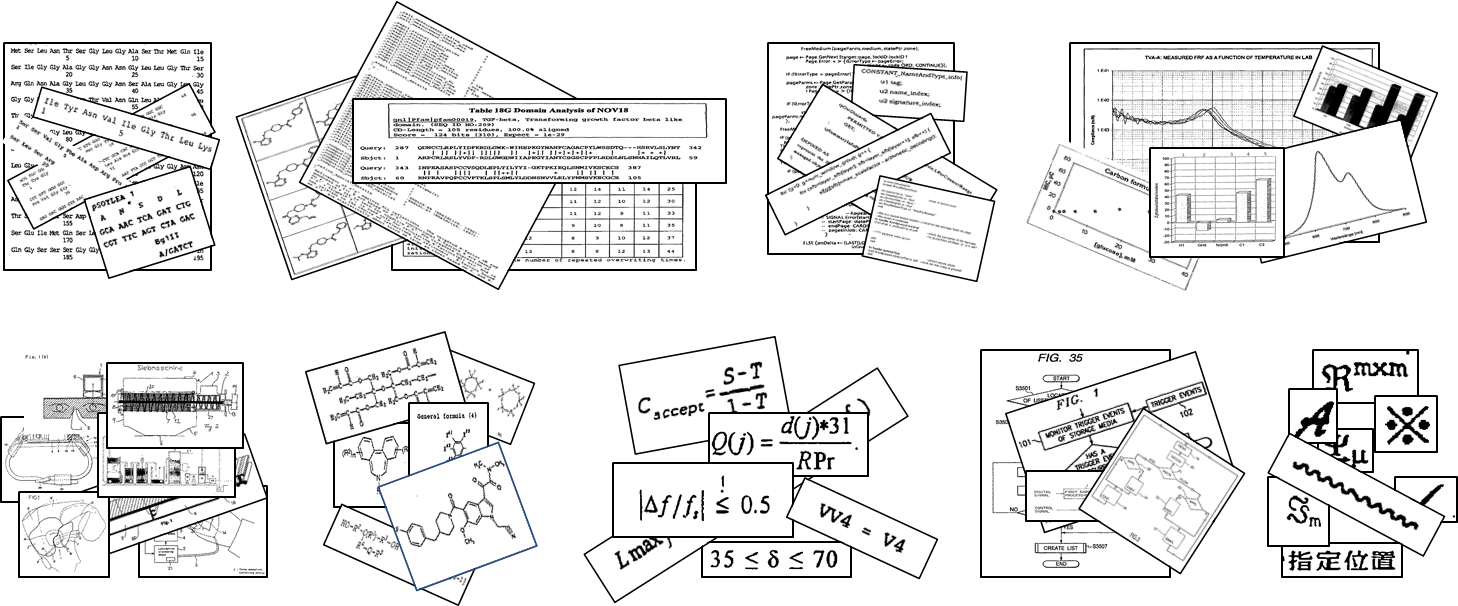}
\caption{Examples from different classes in the CLEF-IP dataset.}
\label{fig.ClefIPEX}
\end{figure}

We used the following document image datasets\footnote{We also considered the 
 NIST Forms dataset~\cite{NIST}, with 20 different classes of tax forms, but 
as the results on this dataset were often of 100\% accuracy, 
these results were not interesting from a parameter comparison study point of view.} 
in our experiments (see examples in Fig.\ref{fig.BYTELEX}-\ref{fig.ClefIPEX} and statistics in Table~\ref{tab:statistics}): 

\textbf{MARG} is the Medical Article Records Ground-truth (MARG) dataset 
\cite{MARG} that consists of 1,553 documents, each document corresponding to the
first pages of medical journals and their size is of 8.4M. 
The dataset is divided in 9 different layout types. Surprisingly the number of columns 
that varied from 1 to 3 within the classes is considered
not relevant to distinguish between classes, which
makes  the dataset challenging  as the "visual" similarity 
is strongly influenced by the number of columns. The criteria of the ground truth labeling is
only the relative
position of the title, authors, affiliation, abstract and the text  
(see for more details \url{http://marg.nlm.nih.gov/gtdefinition.asp}).

\textbf{IH1} is the dataset used in~\cite{Gordo13} that contains 
11,252 scanned documents from 14 different document types (categories) such as  invoices, contracts, IDs,
 coupons, etc.  The images were obtained by scanning paper documents and
their size varies according to the size of the original paper document, 
most of them however having around  3-4M pixels.

\textbf{NIT} is another in-house dataset of 885 multi-page documents with in total 1809 pages  
of 5.6M pixels, but  we only  considered the first page to represent the document\footnote{We did 
experiments with multiple pages where we averaged either the signatures, the similarity scores or 
classification scores,  but using only the first page was most often close to best performance.}
The categories represents as in IH1, document types including   
invoices, mails,  tables, maps, etc, but these documents were not scanned but captured in the print flow
and converted  to images by the print driver (using the Page Description Languages). 
Within this dataset the amount of elements per class varies a lot, 
with several classes having only a few examples, while other classes containing a large percentage of
all documents. We have a second dataset similar to NIT but independent from, that we call \textbf{XRCE} as the documents 
were captured in our own  print flow. This dataset containing mainly scientific articles, patent applications, reports, tables, mails, etc
was used to tune the parameters for some of our parametric models such as SVM or metric learning that were after applied to all the other datasets.

\begin{table}[ttt]
\begin{center}
\bt{|c||c|c||c|c||}
\hline
Dataset & NbIm &  ImSize & NbCls & Example classes\\
\hline
MARG & 1553 & 8.4 & 9 & typeA, typeB \\
IH1   & 11252 & 3-4M & 14 &  invoices, contracts, ID cards,  \\
NIT & 885 &  5.6M & 19  & invoices, mails,  tables, maps \\ 
CLEF-IP  & 38081 & 1.5K -  4.5M & 9 & drawing, graph,  flowchart\\
\hline
\et
\caption{A summary of the dataset statistics}
\label{tab:statistics}
\end{center}
\end{table}

\textbf{CLEF-IP:}   contains the training image set from the 
Patent Image Classification task of the Clef-IP 2011~\cite{clefip11}. 
The  aim of the task in the challenge was to categorize patent images 
into 9 predefined categories  such as 
abstract drawing, graph,  flowchart, gene sequence, program listing, symbol, 
chemical structure, table and mathematics (see examples in \fig{fig.ClefIPEX}). 
The dataset  contains between 300 and 6000 labeled images for each class, 
in total 38081 images with their resolution varying from 1500 pixels to more than 4.5M pixels. 

 \subsection{The experimental setup and evaluation methods}
 \label{sec:setup_eval}
 
We randomly split these datasets   into train (50\%) and test (50\%)  sets five times and 
the same splits are kept along all experiments, 
allowing a comparison between  different features, algorithms  and parameter settings. 

The aim of our experiments is mainly to compare  different image 
representations and to design best practices how to choose the parameters for these representations,
preferably,  independently of the task. Indeed, while the choice of best parameters can be very 
dependent on the task,  with the increasing amount of data it can be more convenient sometimes
to have these features precomputed and pre-stored that allow using the same representations in
various applications such as retrieval, clustering or categorization.
Also, as we already mentioned, for
image type dependent patent search  where the images are first classified,
it can be  practical to use the same features both for the class prediction and for similar image search.

Our  intent therefore is to find feature configurations that perform relatively well across tasks and 
if possible across datasets. Hence, 
we evaluate each representation both in a retrieval framework and using different classifiers (SVM, KNN, NCM)
and we study the behavior of different parameter configurations. 
Note that the Nearest Class Mean (NCM) classifier~\cite{mensinkpami13} that 
predicts the class label of a document image
based on the closest class mean, evaluates  implicitly
(in a certain sense) the ability of these features to 
 perform clustering. Indeed,  NCM, averaging the examples from each  class 
 performs well when these instances  can be easily grouped together. 
 Hence a feature configurations yielding 
 better NCM accuracies is more suitable for clustering 
purposes than one that fails to do it.   Further advantages of the NCM are that it is a multi-class classifier and  
that there is no parameter to be tuned.

When using SVM we used a fixed over all datasets and configurations, which
 means obviously that the SVM results are sub-optimal (in some cases 1-3
  to fine tuned parameters), But in some sense this makes the comparison 
  between  parametric and non-parametric methods such as NCM fairer. Also the focus of the paper is on 
  the parameters of the image representation and fine tuning the parameters of 
  different classifiers or to test more complex classification methods 
  was out of the scope of the paper. To choose the fixed parameter set for the SVM 
  we tested all configurations and a large set of parameters on the XRCE 
  dataset  and considered the setting that performed in general best. As we used one-versus-all linear classifiers with
 stochastic gradient descend~\cite{bottou10compstat} shown to be highly competitive when applied on FVs~\cite{akata14}, 
 the selected parameters were as follows. We used hinge loss with fixed learning rate  $\lambda=1e-5$  for RL features and  
 $\lambda=1e-4$ for FV.  To handle the dataset bias, we weighted the positives by a factor of $\rho=5$ and to optimize the classifier we
  updated the gradient by  passing $N_i=100$ times  randomly through  the whole training set. Similarly, for the same reasons, in the case of the 
KNN classifier we used a fixed $k=4$ as it performed the best on XRCE, but again the results are sub-optimal 
$k=4$ might vary along different datasets and configurations.

To evaluate the classification tasks  using any of the above mentioned classifiers (NCM, KNN and SVM) 
we show only the overall prediction accuracy (OA), \ie the ratio of correctly predicted document images,
but similar behavior was observed when we
considered the average of the per class accuracies\footnote{Or when the behavior was different such as 
for NIT, the reason was that  this dataset contains 
several  classes with only few instances meaning that changing the prediction for any of those  
rare documents may yield a significant change on the accuracy of the corresponding class.}.

To perform document ranking for the retrieval, we use each test document as query and 
the aim is to retrieve  all documents with the same class label in the training set. As similarity between documents we consider 
the dot product between features (which is equivalent with  the cosine similarity as our features are L2 normalized).
To asses the retrieval performance, we use
 mean average  precision (MAP), but we also consider top retrieval accuracy by assessing it 
 by precision at 5 (P@5) as several classes  in these datasets have only few representatives.
 We also consider the precision at 1 (P@1) because this is equivalent with 
the overall classification accuracy of a  KNN classifier using k=1 and hence allows us to 
to compare the results with our actual KNN results for which we used $k=4$.

\section{Experimental Results}
\label{sec:exp}

In this section  we fist do an exhaustive 
parameter study for RunLengths in \mysec{sec:rlexp} and for Fisher Vectors \mysec{sec:fvexp} 
analyzing the behavior of the parameter configurations considering the previously mentioned 
 tasks and datasets. Then, in \mysec{sec:comb_rl_fv}  we discuss about 
possibilities how to merge RL with FV.
 

\begin{table}[t]
\small
\begin{center}
\bt{|c||c|c|c|c||}
\hline
 &  MARG & IH1   & NIT & CLEF-IP   \\
\hline
MAP & \rk{33.8}$\pm  1.4$  & \rk{67.5 }$\pm  0.5$  &  \rk{42.3}$\pm 1.4$   & \rk{38.3}$\pm 0.1$  \\
 & \bk{S5,L4,Q11} & \bk{S5,L1,Q10} &  \bk{S1,L5,Q11} & \bk{S1,L4,Q7} \\
\hline
S & \bk{S5}(96/1)& \bk{S5}(76/.7) & \bk{S1}(96/1) & \bk{S2}(56/3)\\
L & \bk{L5}(80/2) & \bk{L1}(88/1) & \bk{L5}(100/.7) & \bk{L5}(60/2) \\
Q & \bk{Q10}(28/.1) & \bk{Q7}(32/.2)& \bk{Q11}(56/.2) & \bk{Q11}(65/.9)\\
\hline
\hline
KNN & \rk{90.3}$\pm 1.6$  & \rk{95.2}$\pm 0.2$ &  \rk{81.3}$\pm 1.5$ & \rk{84.9}$\pm 0.2$\\
& \bk{S0,L5,Q10}  & \bk{S5,L1,Q9} &  \bk{S5,L5,Q10}  & \bk{S0,L1,Q7} \\
\hline
S & \bk{S5}(60/3) & \bk{S0}(56/.4) & \bk{S5}(58/2) & \bk{S2}(59/1)  \\
L & \bk{L5}(87/5) &  \bk{L1}(74/.7) & \bk{L5}(62/3) & \bk{L1}(100/3) \\
Q & \bk{Q11}(22/.7) & \bk{Q7}(26/.1) & \bk{Q9}(27/.6) & \bk{Q9}(76/2) \\
\hline
\hline
NCM & \rk{61.8}$\pm 1.5$  & \rk{91.3}$\pm 0.3$ &  \rk{66.6}$\pm 0.6$  & \rk{62.4}$\pm 0.2$ \\
 & \bk{S0,L5,Q11}  & \bk{S0,L5,Q11}  &  \bk{S1,L5,Q10}  & \bk{S0,L3,Q11} \\
\hline
S & \bk{S0}(38/3) & \bk{S0}(67/.8) & \bk{S1}(54/2) & \bk{S0}(83/3)  \\
L & \bk{L5}(80/8) &  \bk{L5}(60/4) & \bk{L5}(98/9) & \bk{L4}(50/5) \\
Q & \bk{Q11}(35/1) & \bk{Q11}(48/.5) & \bk{Q10}(34/1) & \bk{Q11}(69/2) \\
\hline
\hline
SVM & \rk{91.9}$\pm  1$ & \rk{96.9}$\pm 0.3 $  & \rk{78.2}$\pm 1.2$ &  \rk{89.8}$\pm 0.2 $\\
& \bk{S0,L5,Q10}  & \bk{S0,L5,Q11}  &  \bk{S5,L5,Q10}  & \bk{S0,L5,Q11} \\
\hline
S & \bk{S0}(35/4)& \bk{S1}(72/.6) & \bk{S5}(35/2.5) & \bk{S0}(71/2) \\
L & \bk{L5}(67/14) & \bk{L5}(65/4) & \bk{L5}(72/5) & \bk{L4}(52/4)\\
Q & \bk{Q11}(31/3) & \bk{Q11}(57/.6) & \bk{Q10}(31/.9) &  \bk{Q11}(66/2) \\
\hline
\et
\caption{Comparative retrieval (MAP) and classification (KNN,NCM,SVM) results where we vary the
parameters  of the  RL features. We show best results in red (averaged over 5 splits) with the corresponding 
configuration in blue (below the accuracy), best parameter frequencies nd performance variations 
best parameter frequencies and performance variations  per feature type.}
\label{tab:QLScomp}
\end{center}
\end{table}

\subsection{Test different parameters for RL}
\label{sec:rlexp}

To build different RunLenght (RL) features, we mainly  varied the image resolution (S), the 
number of layers\footnote{Note that we used 2x2 split of the image at the 
second layer, 4x4 at the third, 6x6 at the fourth and  8x8 at the fifth. Ln means that we concatenated the features of the regions from all the n layers.} 
(L) used in the spatial pyramid and the number of quantization bins (Q).
When we resize an image, we keep the aspect ratio and we define a maximum 
resolution.  We experimented with target resolutions of 50K, 100K, 250K, 500K and 1M pixels and
denoted them  by S1, S2, S3, S4 and S5 respectively. In addition, the case where we do 
not rescale any of the images will be denoted by S0. However, images having less pixels are not upscaled, only images 
above the target size are downscaled. 

In Table \ref{tab:QLScomp}, we show  retrieval performances\footnote{We also assess  P@1 and P@5 
but show only the MAP here.}   ({\bf MAP})  
as well as overall the accuracy of class prediction (OA) for KNN, NCM and SVM
classification.  As we mentioned, the experiments were performed on 5 different splits,
hence in the table we show the mean over the splits and its variation.  
For each dataset and task, in addition to the best average, 
we show, below the best average, the parameter sets that allowed to obtain these results.

In addition, for each parameter type, \eg the number of layer  L,  we alternatively fix
the other parameters, here S and Q, and evaluate the best performing 
value. We do this for all (S,Q) pairs and retain the corresponding variation. 
Then, for each value of the selected parameter type, in this example each Li, we compute the percentage of  time 
 it  performed the best.  In Table \ref{tab:QLScomp} we show for each parameter type the 
 value (\eg Li) that was found the  most often  as best performing. In the parenthesis following the parameter found we  show two numbers. 
 The first  one is the percentage of time that parameter was at the top, the second value shows the average variance of the results 
 for that parameter type (L).  This variation was considered 
by fixing S and Q and evaluating the variance of the results when we varied L, and then averaging over all (S,Q) pairs.
These statistics (frequencies and the variance) were computed by cumulating the results 
along all the 5 splits. Note that if this average 
variance is low, it means that varying 
that parameter has relatively low effect on the obtained accuracy, 
while high average variance means that it is very important to correctly set the given parameter. 
For example, when
we evaluate MARG  with  MAP   we find that L5 performed the 
best 80\% of the time considering all (S,Q) pairs and all splits, and the average variation along L  
when fixing (S,Q) was about 2\%. This means that setting the number of layers is more important
 than the choice of the number of quantization bin, as Q10 was best only 28\% of times and 
 the average variation of Q when fixing 
 (L,S) was only  0.1\%.

\begin{figure}[t]
\centering
\bt{cc}
\includegraphics[width=0.48\textwidth]{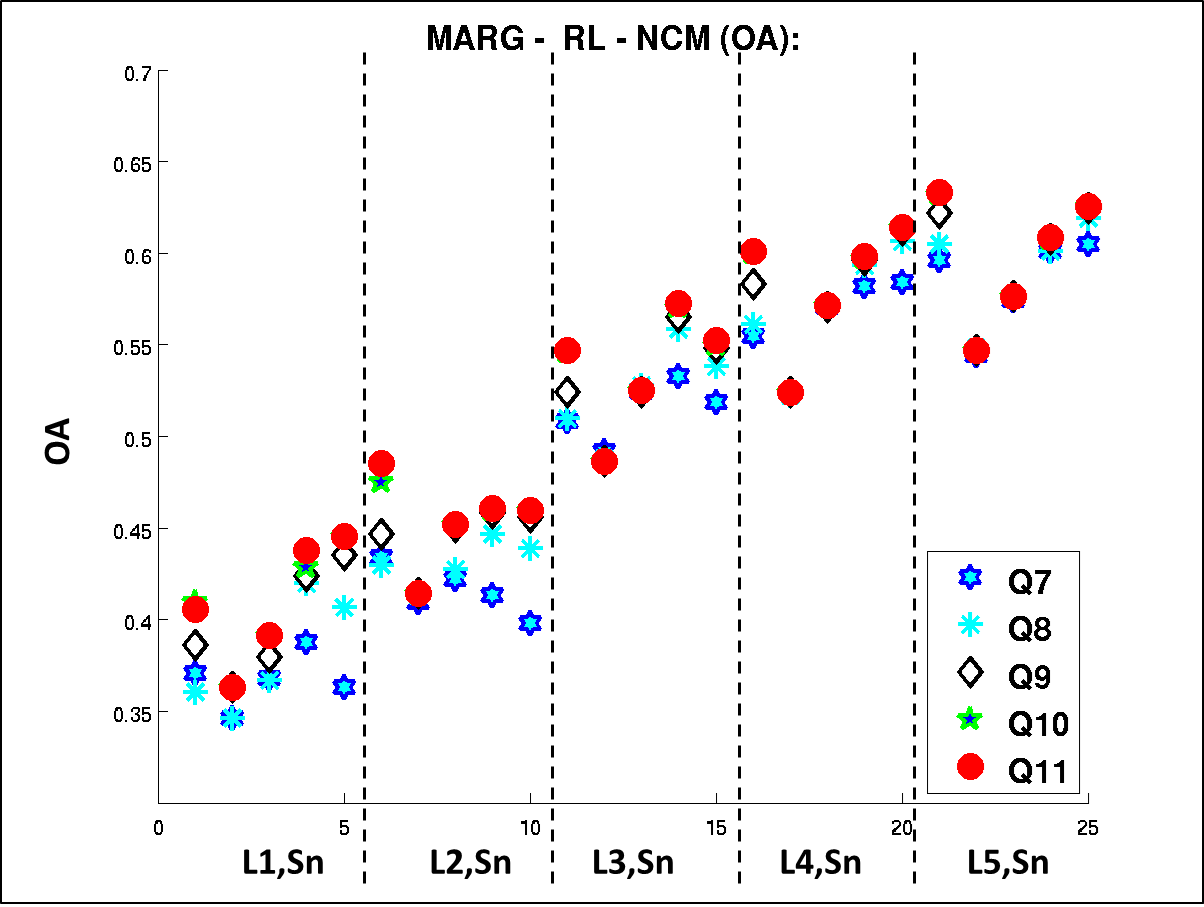} & 
\includegraphics[width=0.48\textwidth]{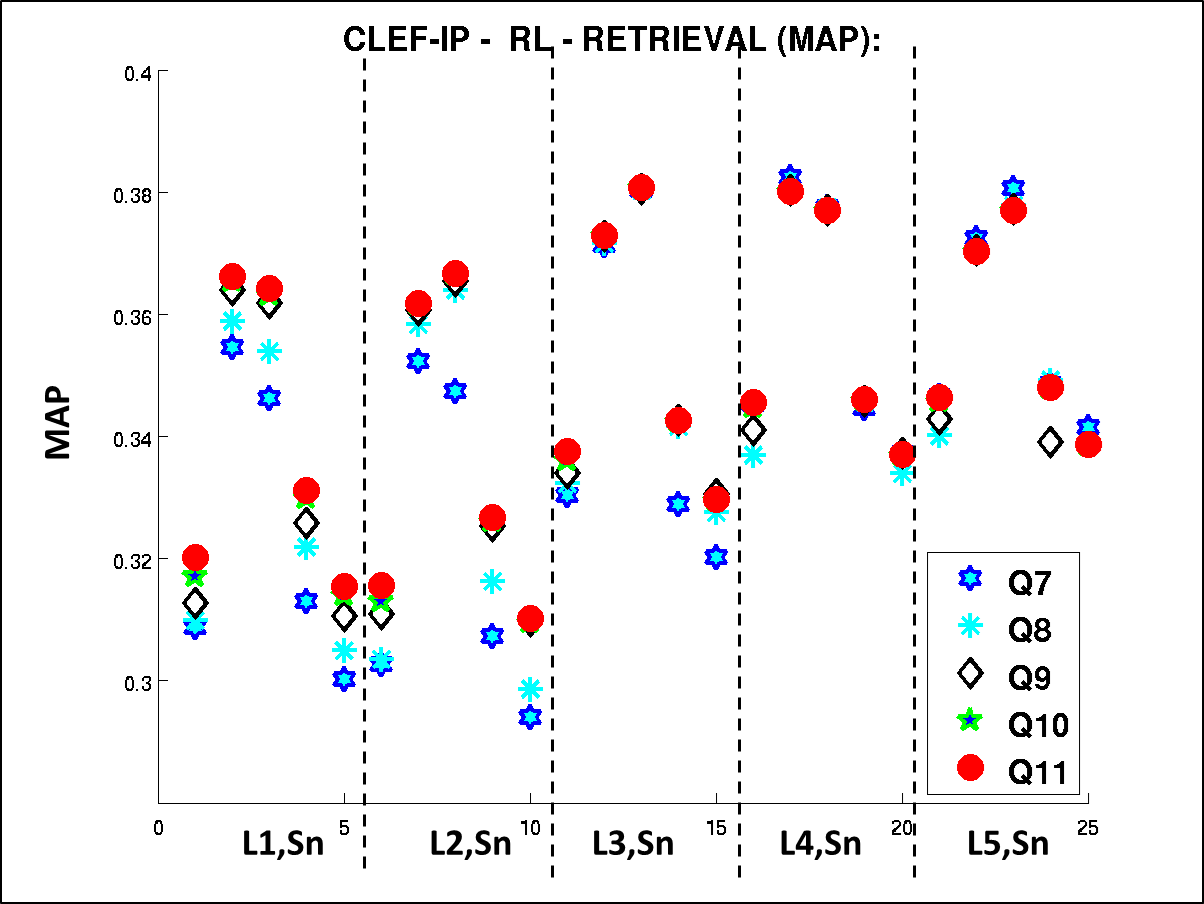} 
\et 
\caption{Example plots comparing different quantizations.}
\label{fig.Qvals}
 \end{figure}

When we analyze the values in Table \ref{tab:QLScomp}, we can deduce the followings: \\

{\bf - Quantization  intervals.}  First of all, we can clearly see that concerning 
 the number of quantization 
intervals, Q11 is almost always the best option. This
shows that considering more quantization values is a good choice. 
On the other hand, the standard deviation and frequency values are relatively low, which means 
that the difference with values obtained using fewer quantization intervals (Q7 to Q10)  
is relatively small,  especially when we have best configurations selected for the other parameters 
(see examples\footnote{All illustrations plot results from the experiments done on the 
first split.} in \fig{fig.Qvals}). \\

\begin{figure}[t]
\centering
\bt{cc}
\includegraphics[width=0.48\textwidth]{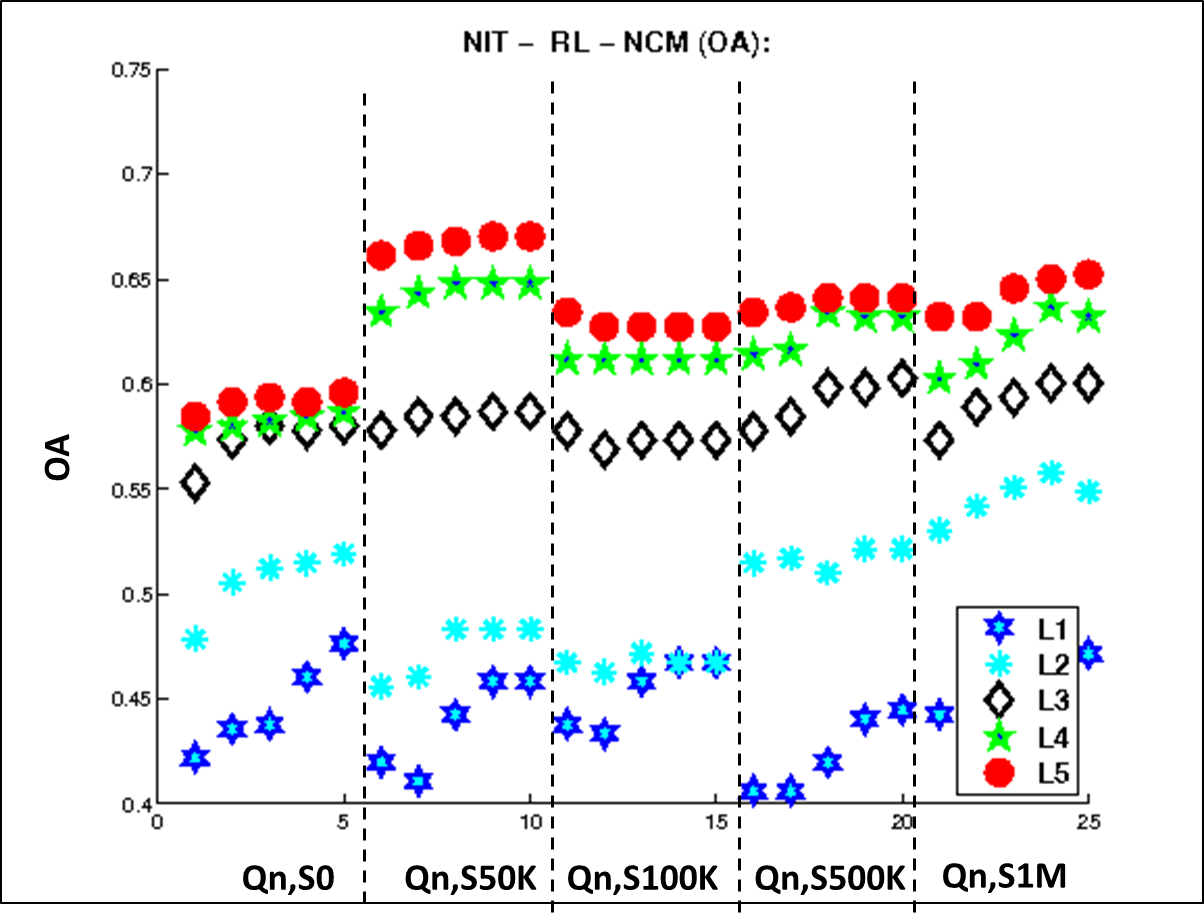} & 
\includegraphics[width=0.48\textwidth]{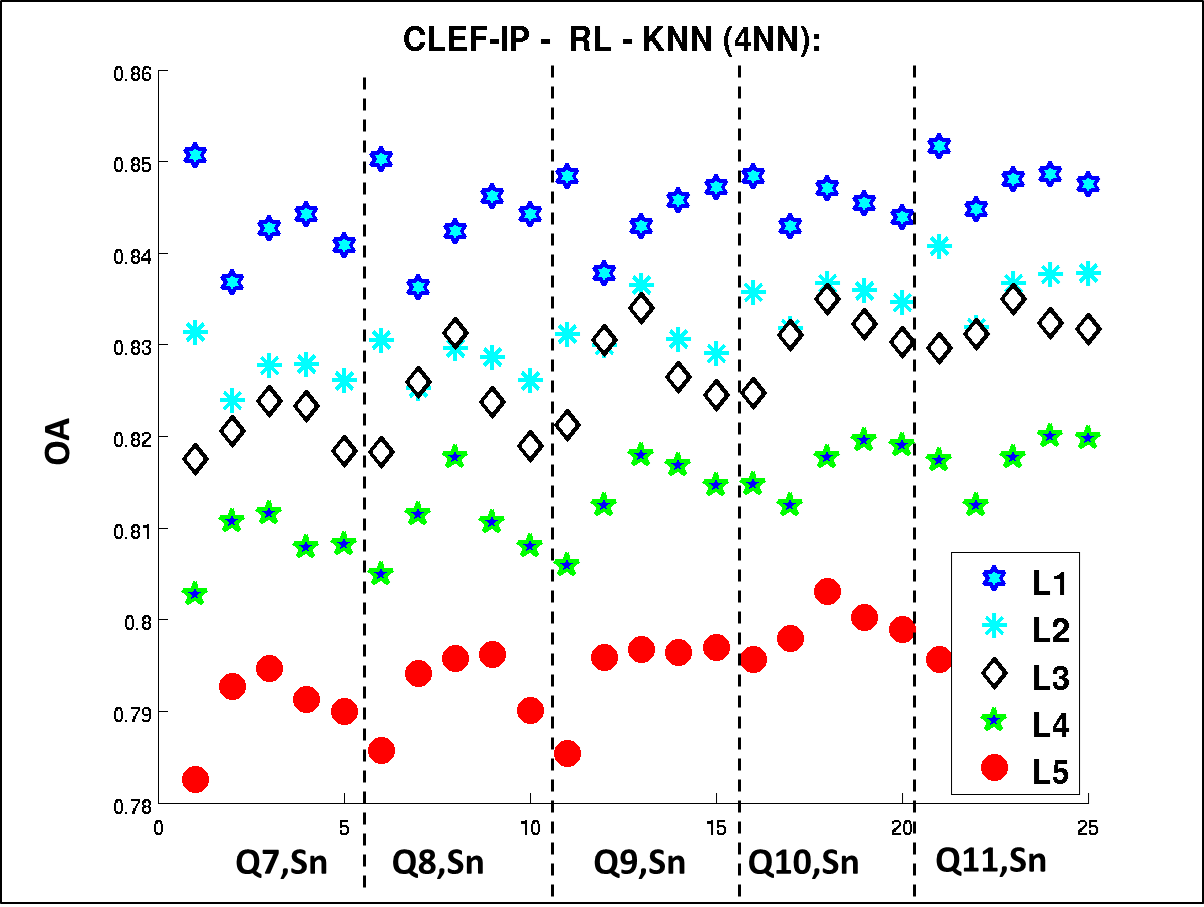}
\et  
\caption{Example images comparing different number  of layers.}
\label{fig.Lvals}
 \end{figure}

{\bf - Number of pyramid layers.} We observe that 
for certain datasets such as  MARG or NIT considering multiple layers (L4 or L5) is essential. This is 
not surprising as the MARG classes   are strongly related to the text  
layout that is much better captured with multiple layers. For other datasets, the 
best layer configuration seems to be 
 task and evaluation measure dependent (see examples in \fig{fig.Lvals}).
Indeed, for IH1 and clefIP, top retrieval results and KNN classification perform
much better using only a single layer, while MAP, NCM and SVM results
 are always better with multiple layers (except the MAP for IH1).  The main reason is 
that  in the former  case, the decision depends only on a few 
"most similar" documents, hence it is sufficient 
to have a few similar documents for most instances in the dataset.  High 
 KNN values ( and top retrieval results, not shown) seems to confirm this for all datasets.

On the contrary, 
the  NCM classifier considers class centroids (\ie averaging over all examples within a class).
Therefore for each test example, it is not any more sufficient the presence of a 
few similar instances but  the similarity 
 to most documents within the class becomes necessary. The MAP evaluates 
how well all instances of a class can be retrieved using an exemplary from the class,
which requires again that the within class similarities to be higher than the similarities
between instances from other classes.  As the NCM and MAP results show, 
these requirements seems to be better 
 satisfied when we consider multiple layers. \\

\begin{figure}[t]
\centering
\bt{cc}
\includegraphics[width=0.48\textwidth]{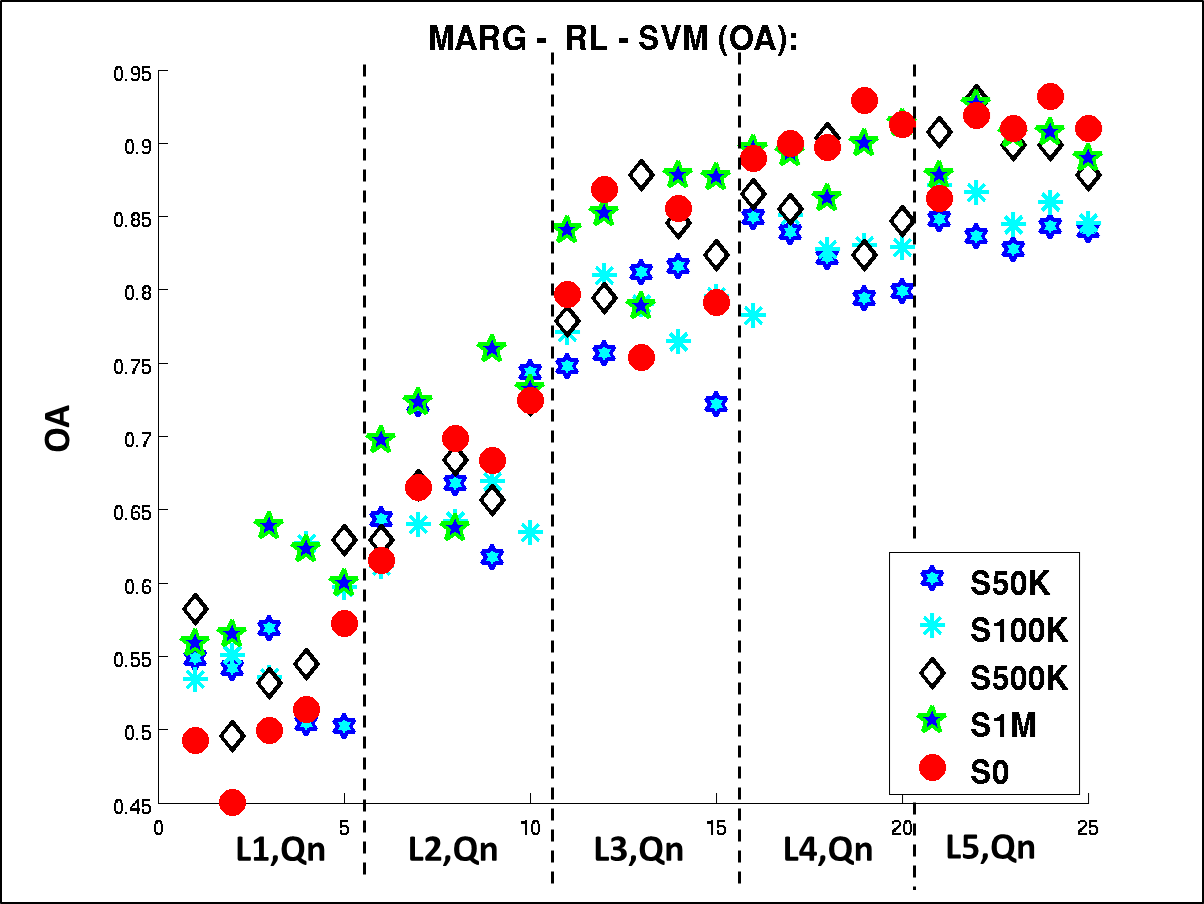} & 
\includegraphics[width=0.48\textwidth]{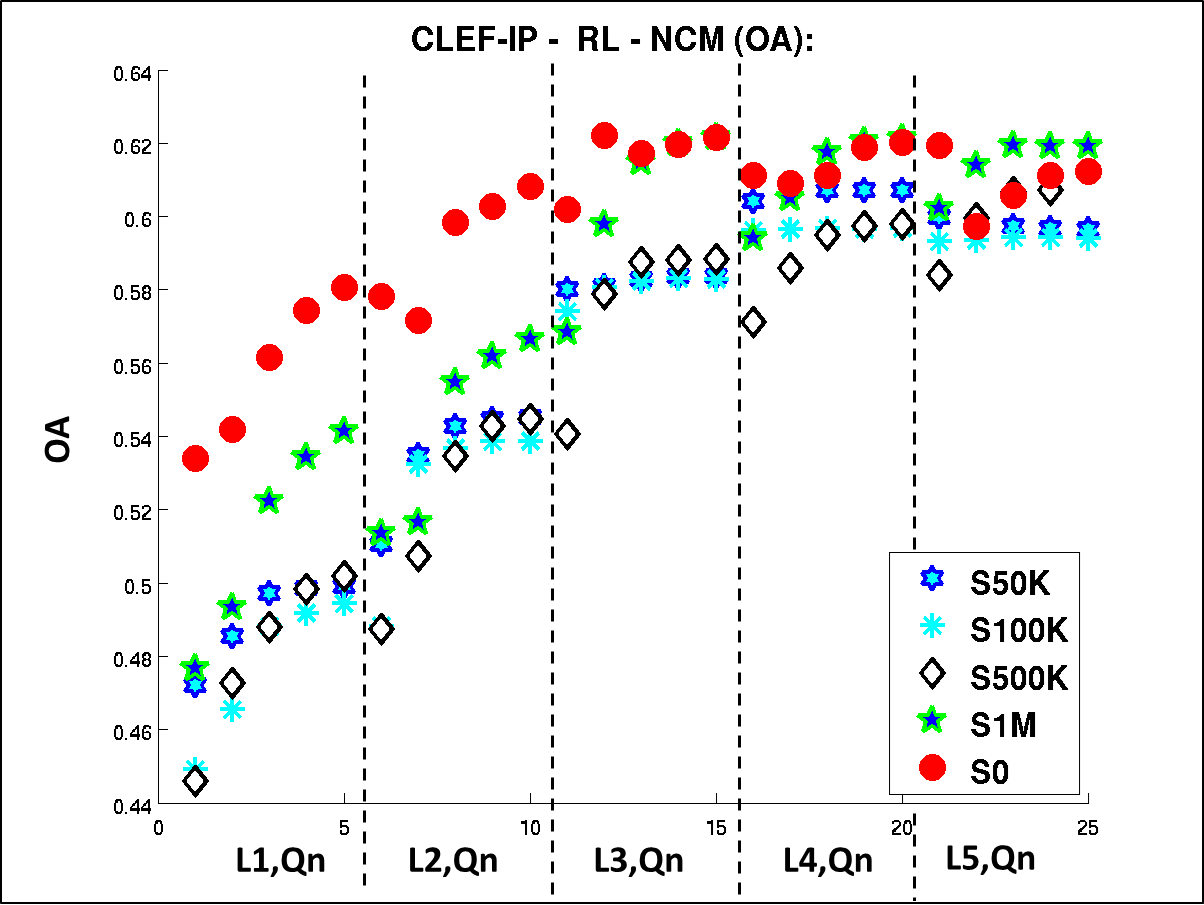} 
\et 
\caption{Examples images comparing  different image resolutions.}
\label{fig.Svals}
 \end{figure}

 {\bf - Image size.} Finally, when we try to observe the effect of  image resizing,
 it is difficult to draw any interesting conclusions.  
Best performing image sizes seem to vary along the datasets, tasks and evaluation measures.  
In a sense, this is not completely surprising, as on one hand 
the original image sizes vary along datasets. Furthermore, while the 
size of the RL does not depend on the image size, the distribution 
of  black and white pixel runs within bins is highly correlated 
with the considered image size. 
Nevertheless, it seems that S0  appears  often as
best performing or yields close to  best performance as we can see in 
\fig{fig.Svals}. The advantage of keeping 
the original size  is to preserve the details present in the image as they were captured,
but on other hand  smoothing can have the benefit  of better generalization. 
Moreover,  for very high resolutional images (which is often the case 
 for document images representing text)  the cost of building the RL vectors from the original   
 images is significantly higher then computing them on  S3 or S4,  especially if we use multi-layer
pyramids. \\

 We now analyze the performances related to different tasks and methods:\\
 
 {\bf - Retrieval.} We can see  that   KNN (and retrieval at top, not shown)
 performs extremely well in general for all datasets, while MAP  performs
 rather poorly. As was 
 discussed above, the good performance obtained with KNN  (and  P@5 not shown) is 
 because for most documents we can find  documents from the same class for which 
 the similarity is high when using well designed RL features. As  we found that in general P@1 
 is higher than  KNN (except for NIT), we can conclude that  using only a single example 
 to classify the documents performs 
 better than using $k=4$ (fixed in our experiments).  On the other hand, 
 the poor  MAP performance   shows  that there is a large  within-class variation
 and it is difficult to retrieve all relevant documents using a single example. Indeed, for example in the 
 case of MARG given a one column document from a class allows to retrieve easily the other 
 one column documents from the same class but has difficulties to rank higher the 
 two column documents from the same class than many of the one column documents from other classes.
However, preliminary results have shown that metric learning approaches 
 specifically designed to support  KNN classification~\cite{Davis07,Weinberger09} or 
 ranking~\cite{Bai09} can significantly improve  the MAP in most cases. \\
 
{\bf - SVM.} The discriminative linear classifier (SVM) even with a set of fixed parameter set
($\lambda=1e-5$, $\rho=5$, $N_i=100$) 
 yields to much better classification accuracy than NCM, showing that
in the corresponding feature space nevertheless the classes are linearly separable. 
In the case of IH1 and clefIP the SVM results are better than nearest neighbor search, 
but for  MARG and NIT they remain below the KNN with $k=1$. Note that
IH1 and clefIP are much larger datasets then MARG and NIT, allowing 
 the SVM to better learn the discriminative classifiers. Moreover, 
the poor SVM results for NIT is not  surprising as 
several classes have very few examples to properly learn the linear classifier in these 
high dimensional spaces. \\

{\bf - NCM.}  Concerning the NCM classification, we made some tentatives to improve the 
NCM by replacing it with NCMC~\cite{mensinkpami13}, but
 the number of optimal centroid per class varies a lot from one class to another and
 fixing the same number of centroids (\eg 2 or 3) does not allow us to significantly 
 improve the classification  performance (except for MARG, where we have clear sub-classes
according to the number of columns). 
KNN with $k=1$ (P@1 results, not shown) is higher than with $k=4$ (except for CLEF-IP) and SVM results 
shows rather good linear separability.  This suggests that the documents within a class are 
not necessary grouped  around a few centroids  but that they are rather scattered in the feature space.

Another way to improve the NCM classification performance is by using 
metric learning as proposed in~\cite{mensinkpami13},
where the distances  between the class means and  the documents are computed 
in the space projected  by a transformation matrix learned on the training set
by maximizing the log-likelihood of the correct NCM predictions in the projective space
(using  a mini-batch stochastic gradient descend (SGD) with 
 fixed learning rate $\lambda=1$  and using 200 random batch of the size equal to the number of classes).
Similarly to SVM, again we trained the NCM metric learning approach with the parameters below  obtained as 
best on the XRCE dataset. This means that the results are suboptimal and we could improve the results further by
tuning them on each tested datasets, which could be another option that was out of the scope of this paper. 

Concerning the learned projected space, we experimented with different target 
dimensions  $D$ such as 16, 32, 64 and 128 and in  Table \ref{tab:pca} we show the best results obtained with 
the  corresponding parameters.  In addition, we also present
the results when the projection was made using the $D$ first principal 
directions\footnote{Note that we initialize the metric learning with PCA.} 
of the PCA.  In this set of experiments we only varied the number of layers in the pyramid, 
did  no image rescaling (S0) and used  $Q=11$ quantization bins. We can see that while using PCA the 
results remain almost the same as without, the metric learning allows us to significantly 
improve the NCM performance in all cases. This time the NCN performance is 
close to the  KNN and SVM performances, and could be further improved if we
 fine tune the parameters on the tested dataset. \\

\begin{table}[t]
\small
\begin{center}
\bt{|c||c|c|c|c||}
\hline
$\bm W$ &  MARG & IH1   & NIT & CLEF-IP   \\
\hline
$I$ & \bk{61.8}(L5,D0648)   & \bk{91.3}(L5,D10648)  &  \bk{59.8}(L5,D10648)   & \bk{62.4}(L3,D1848)\\
PCA & \bk{61.4}(L5,D128)   & \bk{91.3}(L5,D128)  & \bk{59.6}(L5,D128)   & \bk{62.1}(L3,D128)\\
ML & \rk{89}(L5,D64) & \rk{92.6}(L5,D64) & \rk{81.3}(L5,Q11) & \rk{86.6}(L5,D128)\\
\hline
\et
\caption{Comparison of the NCM results without projection ($\bm W=\bm I$), with PCA projections and 
 with metric learning.}
\label{tab:pca}
\end{center}
\end{table}

Note that the projected features are much smaller that the original 
features, especially for the multi-layer pyramid where for Q11 and L5 we go from 
10648 to 128 dimensions, which can be interesting in case we want to store the features. 
We also experimented KNN and SVM with the PCA reduced features 
and we observed that, similarly to the NCM results,  we were able to keep  
similar performances in all cases in spite of the strong dimension  reduction. 
While similar observation was made in \cite{Gordo13}, 
 Gordo \etal \cite{GPV13}  proposes a compression and 
 binarization  through PCA embedding that significantly outperforms the results 
 obtained with simple PCA.


\subsection{Test different parameters for FV}
\label{sec:fvexp}

Similarly to the previous section, we tested different parameter configurations 
for the Fisher Vectors (FV) built on local SIFT descriptors. To build the FV in natural images,
first local patches (windows of size NxN) are extracted densely at multiple scales 
and SIFT descriptors are computed on each of them. For $N$  we consider the values
 24, 32, 48 and 64 and denote them by W1, W2, W3 and W4 respectively.

In the case of document 
images, especially when the images are of similar resolution, 
extracting features at multiple scales might have less importance, therefore in the  
first set of experiments we focus
on extracting features only at a single scale. Also, as
the resolution of most  original images is very large and the considered local patches 
 relatively small, we first resize the images\footnote{When using FV with  natural images we often 
resize the images first,  often to 100K pixels~\cite{Perronnin07,PSM10}.} to have a maximum of 250K pixels (S3). 
Then on each local window, we compute the usual 128 dimensional SIFT features~\cite{lowe04ijcv}, 
and reduce them to  48, 64 and 96 dimensions using PCA. We
denote the corresponding low level features by F1, F2 and F3. 
While we can also build FV with the original SIFT features, we do not report results on it as 
we observed that reducing the dimensionality not only  significantly decreases
 the size of the FV,  but in general the accuracy is also improved.

\begin{table}[ttt]
\small
\begin{center}
\bt{|c||c|c|c|c||}
\hline
 &  MARG & IH1   & NIT & CLEF-IP   \\
\hline
MAP &  \rk{34.2}$\pm 0.3$ & \rk{73.5}$\pm 0.2$  &  \rk{44.6}$\pm 1.5$ & \rk{41.9}$\pm  0.1$ \\
 & \bk{W3,F1,G4} & \bk{W3,F2,G1}  &  \bk{W3,F2,G1} & \bk{W4,F1,G1}\\
\hline
W & \bk{W3}(52/2) & \bk{W4}(71/9) & \bk{W4}(67/.8) & \bk{W3}(42/4)\\
F & \bk{F1}(80/.8) & \bk{F1}(69/2)& \bk{F1}(61/.5) & \bk{F1}(75/2)\\
G & \bk{G1}(35/1) & \bk{G1}(70/6) & \bk{G1}(68/1) &  \bk{G1}(98/9) \\
\hline
\hline
KNN & \rk{89.8}$\pm 1.3$ & \rk{95.4}$\pm 0.3$  &  \rk{82.1 }$\pm 2$ & \rk{86.2}$\pm 1.1$  \\
 & \bk{W3,F1,G4}  & \bk{W4,F1,G3}  & \bk{W2,F1,G3} & \bk{W3,F1,G2}\\
\hline
W & \bk{W3}(53/11) & \bk{W4}(100/6) & \bk{W2}(35/2) &   \bk{W3}(41/9) \\
F & \bk{F1}(71/4) & \bk{F1}(84/2) & \bk{F1}(36/.8) & \bk{F1}(76/5)\\
G & \bk{G1}(32/6) & \bk{G1}(77/5) & \bk{G5}(37/1) & \bk{G1}(89/17)\\
\hline
\hline
NCM &  \rk{69.1}$\pm 0.9$ & \rk{91.6}$\pm 0.3$ &  \rk{77.8}$\pm 3$ & \rk{69.9}$\pm 0.1$\\
& \bk{W3,F1,G5} & \bk{W3,F1,G4}  & \bk{W1,F3,G7}& \bk{W3,F1,G1} \\
\hline
W & \bk{W3}(95/4) & \bk{W3}(78/6) & \bk{W2}(55/7)&  \bk{W3}(95/4) \\
F & \bk{F2}(37/1) & \bk{F1}(54/.3) & \bk{F3}(53/1) & \bk{F1}(64/2)\\
G & \bk{G7}(57/3) &  \bk{G4}(37/.4)& \bk{G7}(88/6) &  \bk{G1}(58/3)\\
\hline
\hline
SVM & \rk{87.4}$\pm 1 $  & \rk{97.3}$\pm 0.2 $  & \rk{85.7}$\pm 2.3$ & \rk{87.6}$\pm 0.2 $ \\
& \bk{W4,F1,G4}  & \bk{W3,F3,G6}  & \bk{W2,F2,G6} &  \bk{W3,F3,G6} \\
\hline
W & \bk{W4}(86/4) & \bk{W3}(79/.3) &  \bk{W3}(42/1) &   \bk{W3}(100/2) \\
F & \bk{F1}(48/1) & \bk{F3}(54/.2) & \bk{F2}(37/1)& \bk{F3}(88/.8) \\
G & \bk{G6}(25/3) & \bk{G5}(55/.4) & \bk{G6}(27/4) & \bk{G7}(70/1) \\
\hline
\et
\caption{Comparative retrieval (MAP) and classification (KNN,NCM, SVM) results where we vary 
the size of the local window  (W), the dimension of the PCA-reduced SIFT features (F) and 
the number of Gaussians used in the visual vocabulary (G) to build the FVs. 
We show best results in red (averaged over 5 splits) with the corresponding 
configuration in blue (below the accuracy), best parameter frequencies and performance variations 
per feature type.}
\label{tab:WFD4comp}
\end{center}
\end{table}

In a given projected feature space, \eg corresponding to W3 and F2,
we build a set of visual vocabularies using Gaussian Mixture Model (GMM)
 with diagonal covariance matrices, where we vary the number of visual words by
considering $2^{(g+3)}$ Gaussians where $g=1..7$. We  denote the corresponding 
vocabularies by G1,..G7, where \eg G1 corresponds to 16 Gaussians and G5 to 256 Gaussians.
Note that both the PCA projection matrices and the GMM models 
were built using the features extracted on the XRCE dataset and then applied to all the other datasets.
This  means that  for a given parameter setting (W,F,G) the documents are 
represented exactly in the same  feature space (FV) independently 
of the dataset on which the experiments are done\footnote{At the end of this section, 
to show the influence of the visual model, we provide a few results with
the visual vocabulary built on the same dataset on which the experiments
were performed.}.

In this first set of experiments, as we do not use any spatial pyramid, we 
have only three varying parameters  for the FV: the size of the local
window  (W), the dimension of the PCA-reduced SIFT features (F) and 
the number of Gaussians used in the visual vocabulary (G).  
Retrieval (P@1, P@5 and MAP) and overall (OA) classification (with  KNN, NCM and SVM) 
 accuracies with the best parameter settings, winning frequencies and variances
are shown in Table \ref{tab:WFD4comp}.  From these results
we can conclude the followings:\\

{\bf - Feature size:} Best results are obtained  in general 
with F1 (SIFT reduced to 48 dimensions)  or, when it is not the case (\eg NCM applied to NIT or 
SVM applied to CLEF-IP), the low average variances suggest that the corresponding 
results obtained with F1 are not very different.  \\

 \begin{figure}[t]
\centering
\bt{cc}
\includegraphics[width=0.45\textwidth]{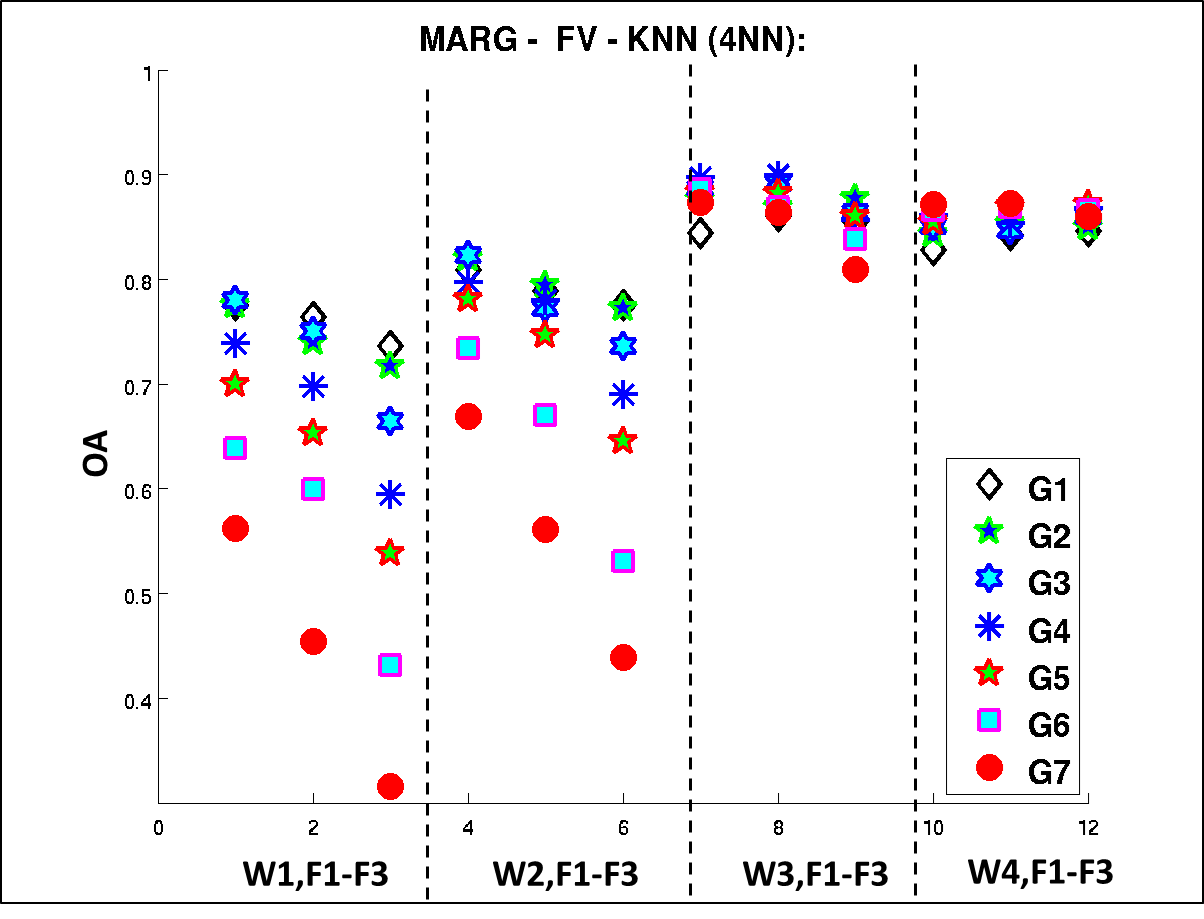} &
\includegraphics[width=0.45\textwidth]{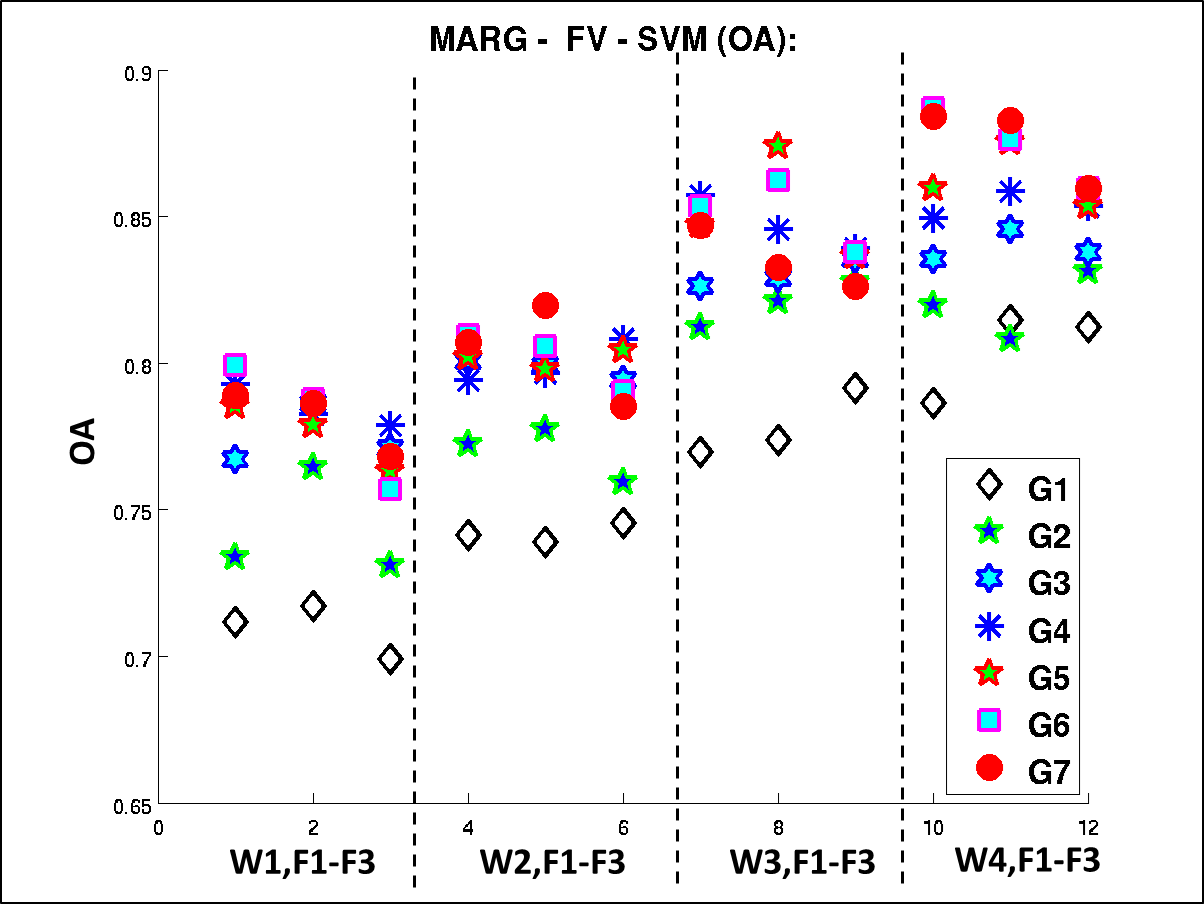} \\
\includegraphics[width=0.45\textwidth]{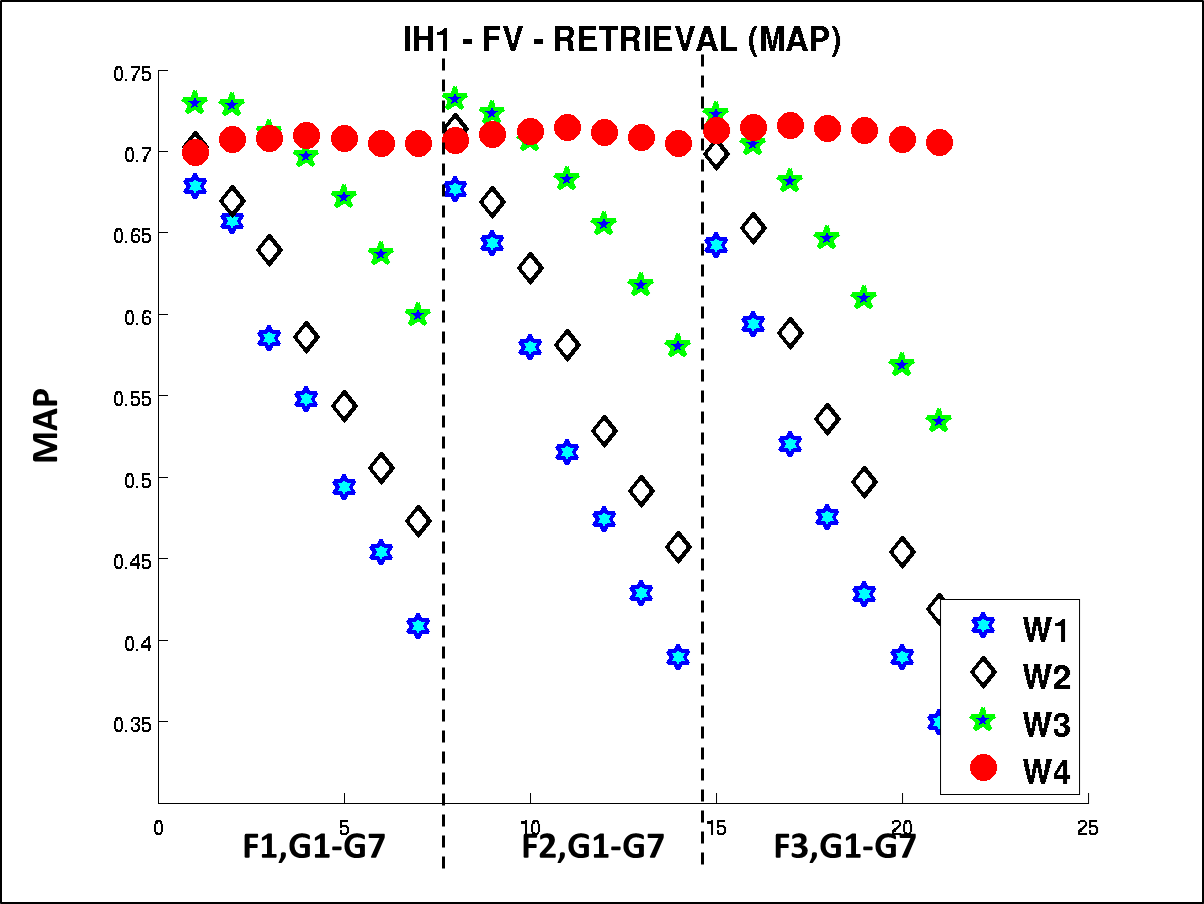} & 
\includegraphics[width=0.45\textwidth]{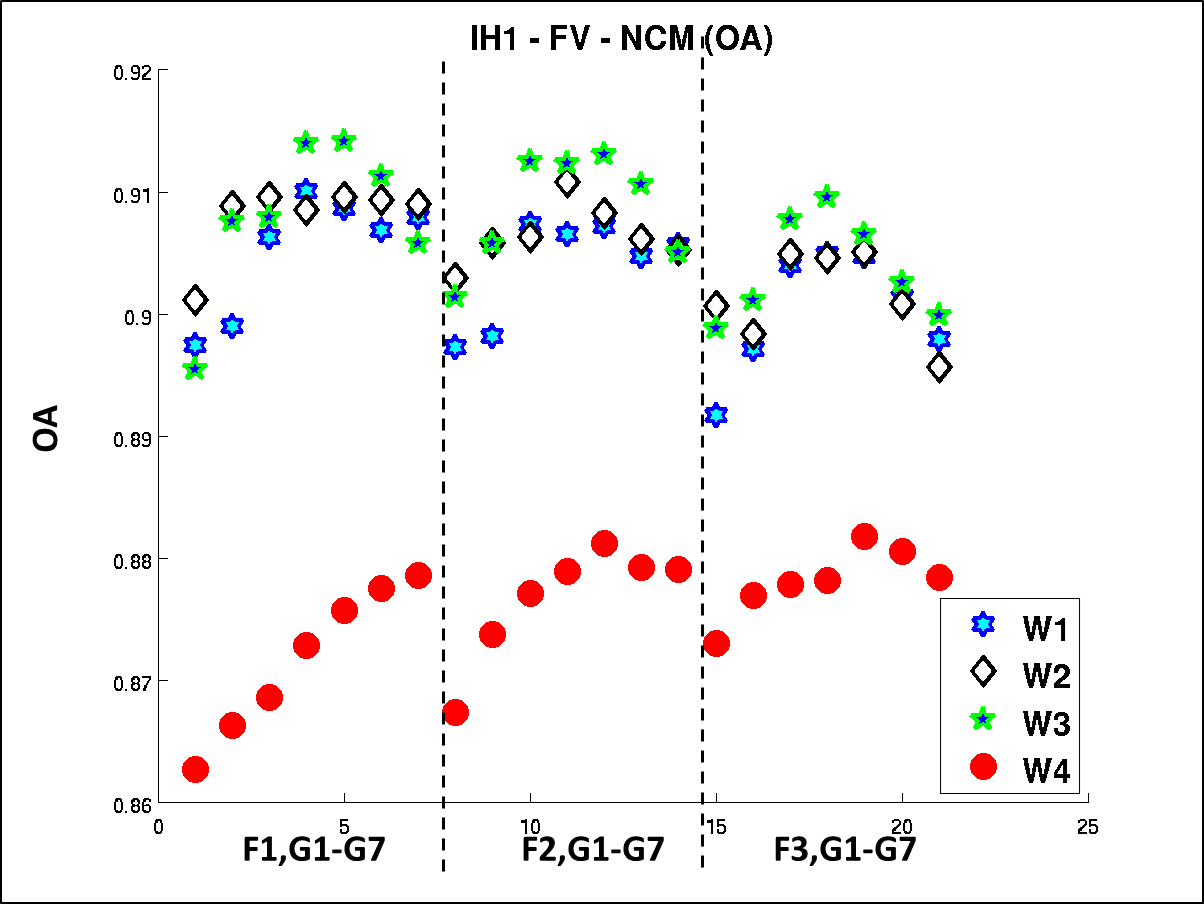} 
\et 
\caption{Example plots comparing different vocabulary sizes (top) and window sizes (bottom).}
\label{fig.GWcomp}
 \end{figure}

 {\bf - Vocabulary size.}  The number of Gaussians seems to be dependent both on the dataset 
 and the tasks.  In the case of retrieval, best results are obtained  with smaller vocabulary sizes,
and hence much smaller image signatures, most often G1 (except top retrieval on NIT). 
On the other hand,  NCM and SVM seems to require  much larger vocabularies, 
which is not surprising especially concerning SVM.  We observe that,
while  extreme values (G1 or G7) yield often best or worse results, using
values between 64 (G3) and 256 (G5)  seems to be a good compromise between size and accuracy
(see also \fig{fig.GWcomp}, top row).\\

 {\bf - Window size:} W3 (patches of size 48x48) seems to be  often best performing or a good compromise 
 compared to the other cases. Indeed, W1 performs in general poorly showing that a  too small window 
 (containing rather few information about the page) is not a good idea. While W4 can perform 
 best for certain tasks (\eg retrieval on Bytel) it can be worse on another task (NCM on Bytel),
 as shown  in \fig{fig.GWcomp}, bottom row. The variances are also high, showing the importance of 
 setting this parameter properly.   \\
 
 Note nevertheless that the ideal 
size of the window is strongly correlated with the processed image size. 
If we increase the image size
we need  larger windows to capture the same amount of information. 
Decreasing the image while keeping the same window size allows to increase the amount of 
information per windows. Another possibility to ensure we capture the information at 
the right scale is to extract the features at multiple scale. Therefore in which follows we vary both the image size and
the number of scales at which the features are extracting,  while fixing the window size to 48x48 (W3). 

As we observed that S1 (50K) performs in general poorly, we consider the following image sizes:
100K (S2), 250K (S3) and 500K (S4) pixels respectively.  Note that we do not consider  
S0 and S5 (1M pixels) as they lead not only to extremely 
large amount of windows (increasing significantly the computational cost),  
but also the information captured within such  window  is
 extremely poor in content (many of them containing only white pixels). 
As we have seen that W1 on S3 (containing much less information than W3) performed poorly, 
we expect even worse results when using W3 with S0 or S5.  To get similar or possible better results
one might consider much larger windows to compute the SIFT features. 

To handle the feature extraction at multiple scale
we further downscale the image (\eg S3) by a scale factor of $\sqrt{2}$ and extract SIFT 
again from  windows of size 48x48. 
The amount of information in these windows extracted from the image downscaled by $\sqrt{2}$ 
corresponds to the (smoothed) information extracted from S3 with a window W3 
upscaled by $\sqrt{2}$. We repeat this process until which we reach the number
of desired scales. We experimented with 1,3,5 and 7  scales
 denoting them by M1, M3, M5 and M7. Note that in the case of the configuration 
 (S3,W3,M5) this means that the images of size S3 were 5 times downscaled by $\sqrt{2}$
 and at each scale the SIFT features were extracted on windows of size 48x48. These features are PCA-reduced to a dimension of 48 (F1) 
 and  all cumulated to form the feature set $X_I$ that generates the  FV corresponding to the image.

\begin{table}[ttt]
\small
\begin{center}
\bt{|c||c|c|c|c||}
\hline
{\bf RET} &  MARG & IH1   & NIT & CLEF-IP   \\
\hline
MAP &  \rk{35.3}$\pm 0.1$ & \rk{75.7}$\pm 0.2$  &  \rk{46.6}$\pm 1.5$ & \rk{46.1}$\pm  0.8$ \\
 & \bk{S3,M5,G5} & \bk{S3,M3,G2}  &  \bk{S3,M5,G1} & \bk{S2,M7,G2}\\
\hline
S & \bk{S3}(64/.9) & \bk{S3}(68/5) & \bk{S3}(89/1) & \bk{S2}(69/2)\\
M & \bk{M5}(76/.9) & \bk{M7}(57/4)& \bk{M5}(81/.7) & \bk{M7}(88/3)\\
G & \bk{G5}(63/.9) & \bk{G1}(65/6) & \bk{G1}(83/2) &  \bk{G1}(78/3) \\
\hline
\hline
KNN & \rk{91.4}$\pm .9$ & \rk{95}$\pm 0.2$  &  \rk{81.7 }$\pm 2$ & \rk{89.9}$\pm 0.8$  \\
 & \bk{S3,M7,G3}  & \bk{S3,M3,G1}  & \bk{S4,M7,G5} & \bk{S4,M7,G2}\\
\hline
S & \bk{S3}(81/3) & \bk{S3}(87/2) & \bk{S4}(68/2) &   \bk{S4}(94/4) \\
M & \bk{M5}(57/2) & \bk{M5}(44/1) & \bk{M5}(38/.9) & \bk{M7}(43/4)\\
G & \bk{G4}(33/2) & \bk{G1}(85/4) & \bk{G3}(25/1) & \bk{G1}(49/5)\\
\hline
\hline
NCM &  \rk{71.2}$\pm 1.6$ & \rk{92.2}$\pm 0.2$ &  \rk{75.2}$\pm 2.1$ & \rk{75.9}$\pm 0.3$\\
& \bk{S3,M7,G6} & \bk{S3,M3,G5}  & \bk{S4,M3,G7}& \bk{S2,M5,G5} \\
\hline
S & \bk{S3}(59/2) & \bk{S3}(53/.9) & \bk{S4}(43/3)&  \bk{S2}(86/2) \\
M & \bk{M5}(60/2) & \bk{M3}(44/.5) & \bk{M1}(81/2) & \bk{M5}(49/3)\\
G & \bk{G5}(58/4) & \bk{G4}(42/.7)& \bk{G7}(58/4) &  \bk{G1}(27/1)\\
\hline
\hline
SVM & \rk{91}$\pm 1.1 $   & \rk{97.4}$\pm 0.1 $  & \rk{86.5}$\pm 2.1$ & \rk{94.7}$\pm 1.1$ \\
& \bk{S4,M7,G5}  & \bk{S4,M7,G7}  & \bk{S3,M3,G4} &  \bk{S4,M7,G7} \\
\hline
S & \bk{S3}(57/2) & \bk{S3}(64/.4) & \bk{S3}(64/1) &   \bk{S4}(93/1) \\
M & \bk{M5}(45/2) & \bk{M5}(63/.3) & \bk{M3}(38/.6)& \bk{M5}(57/2) \\
G & \bk{G5}(58/3) & \bk{G7}(51/.4) & \bk{G5}(50/1) & \bk{G7}(88/1) \\
\hline
\et
\caption{Comparative FV results on different datasets and tasks. 
We show best results in red  and best parameter 
settings in blue using different evaluation measures.}
\label{tab:SMGcomp}
\end{center}
\end{table}

 The result with varying number of scales (M), image resolutions (S) 
and vocabulary sizes (G) are shown in Table \ref{tab:SMGcomp}. 
From these results, we can conclude the followings: \\

\begin{figure}[t]
\centering
\bt{cc}
\includegraphics[width=0.45\textwidth]{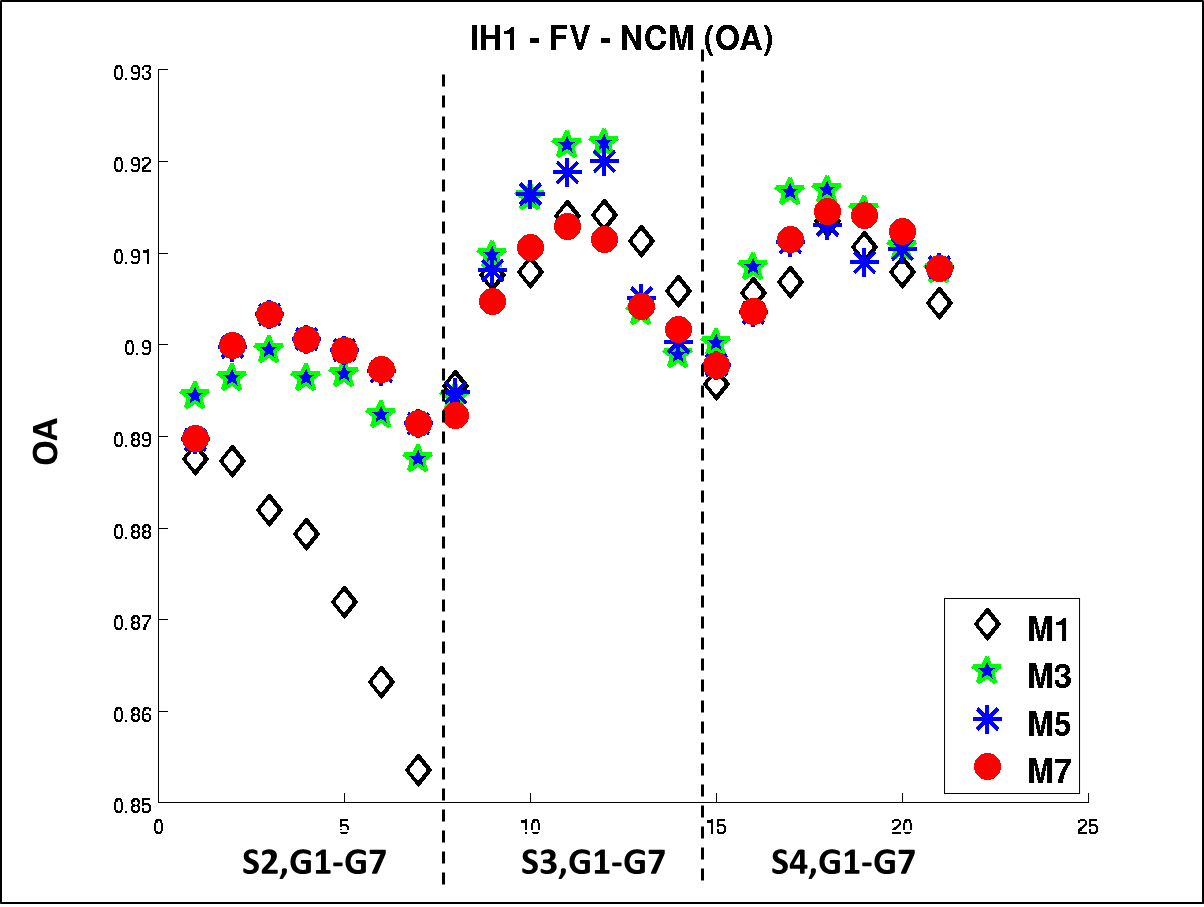} & 
\includegraphics[width=0.45\textwidth]{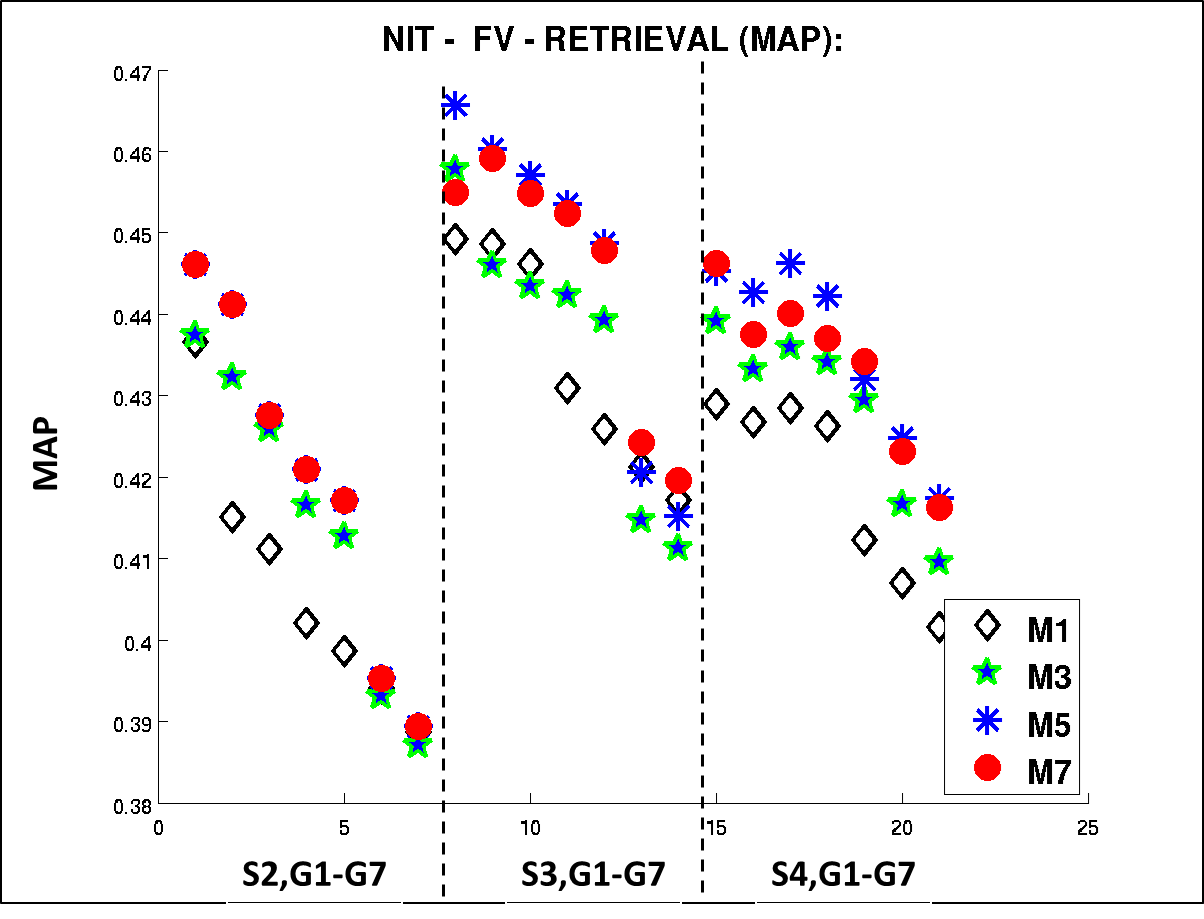} \\
\includegraphics[width=0.45\textwidth]{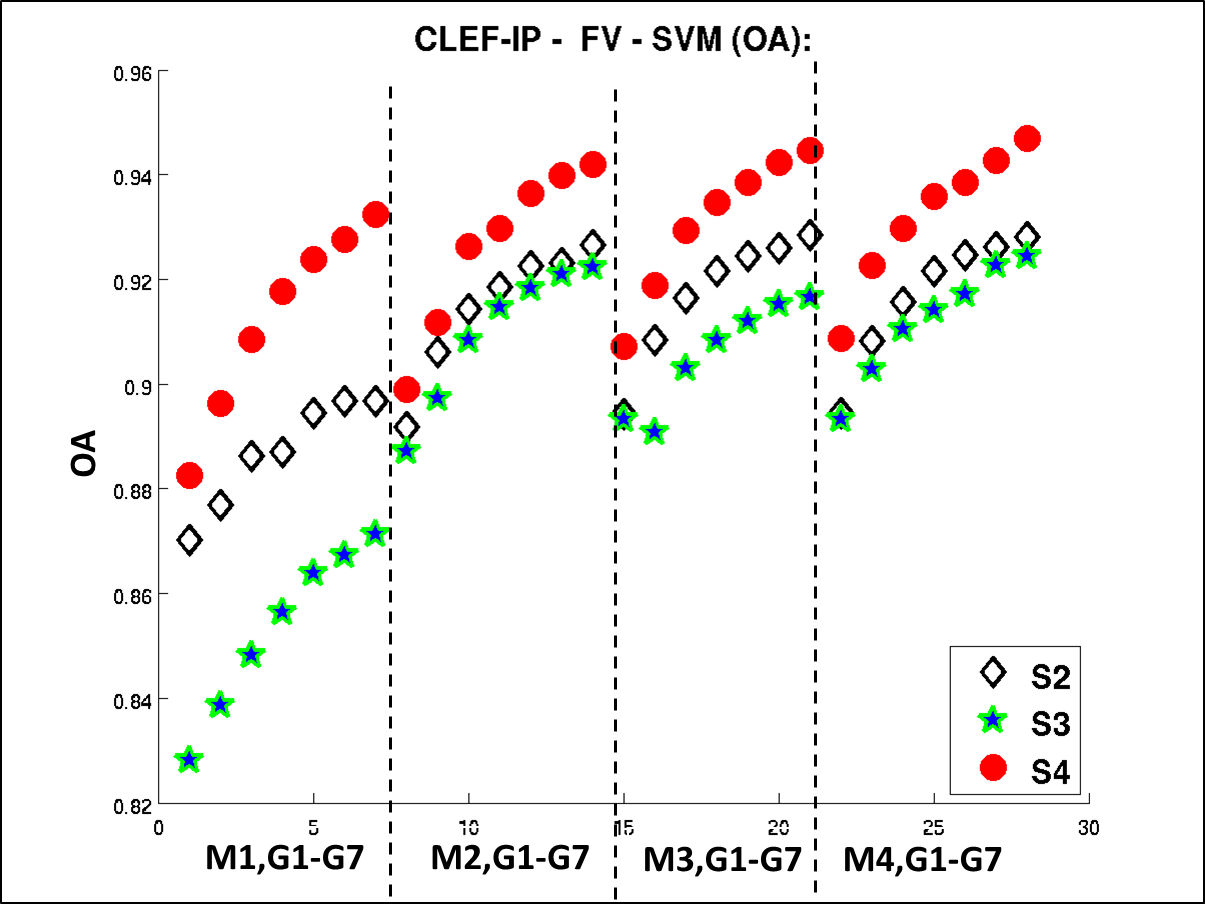} & 
\includegraphics[width=0.45\textwidth]{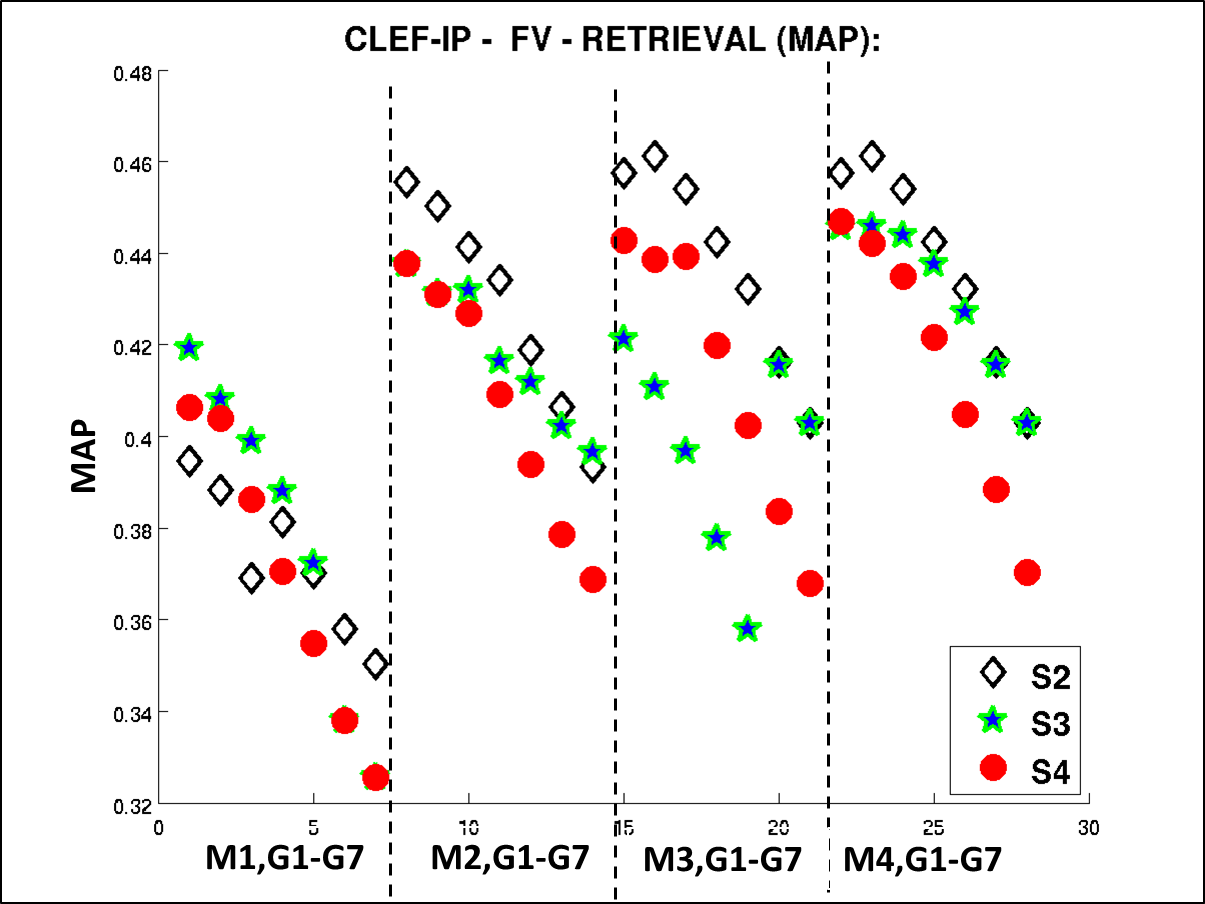} 
\et 
\caption{Example plots comparing different number of scales (top) and image sizes (bottom).}
\label{fig.MScomp}
 \end{figure}

{\bf - Number of scales.} Concerning the number of scales (M), 
extracting  features at multiple levels definitely helps. While 
there is no clear winner between M3, M5 and M7,  as the plots in  \fig{fig.MScomp} top row shows 
they have similar performances and in general all outperform M1. This suggest that 
while it is important to consider multiple scales,  considering
3 or 5 scales, are in general, sufficient. \\

{\bf - Vocabulary size.} The behavior of the visual vocabulary size remains similar to 
our previous set of experiments where we varied the window size (W) and the  feature dimension (F).
Again, while extreme values (G1 or G7) are often best or worst,  G3, G4 and G5 are often close 
to best or even  winning  in the case of MARG where G5 performs the best on all tasks. \\

{\bf - Image size.} 
S3 (250K pixels) is the best performing in most cases with MARG and IH1, showing that the 
configuration (S3,W3) is  suitable for them. Concerning NIT and CLEF-IP, there is no clear winner
(see also \fig{fig.MScomp}, bottom rows). For nearest neighbor search based methods
(KNN, P@1,P@5) and SVM when using window size of W3 it seems better to keep 
higher resolution (S4) while 
retrieval  and NCM classification on  CLEF-IP works better with low resolution (S2). \\

Finally,  we test the spatial pyramid with several layers for the FV features, where
we consider the  number of maximum layer in the pyramid in function of the visual vocabulary as
FV features being already very large for vocabulary sizes above 64  to obtain
the final signature  size they are multiplied by the  number of regions in the pyramid. 
We show in Table \ref{tab:Gstatistics} the maximum 
number of layers we consider for each  vocabulary size in our experiments. We also show the 
number of corresponding regions and  the size of the final signature 
(after the concatenation of the FVs for all the regions).

\begin{table}[ttt]
\small
\begin{center}
\bt{|c||c|c|c|c|c|c|c||}
\hline
G & G1 & G2 & G3 & G4 & G5 & G6 & G7 \\ 
\hline
L & 5 & 5 & 4 & 3 & 3 & 2 & 2 \\
\hline
Nb reg &  121 & 121 & 57 & 21 & 21 & 5 & 5\\
\hline
FD  & 185856 & 371712 & 350208 & 258048 & 516096  & 245760 & 491520 \\
\hline
\et
\caption{A summary of the feature sizes}
\label{tab:Gstatistics}
\end{center}
\end{table}

We can see that these signatures are very large and in general not sparse and we did all
our experiments with non-compressed FVs. Note nevertheless that there are several methods in the literature 
\cite{perronnin10cvpr,sanchez11,GordoCVPR11,vedaldiaz12}
that propose to efficiently binarize and/or
compress the Fisher Vectors while keeping them highly competitive. It could be interesting, 
but  testing the effect of  methods in the case of different configurations
was out of the scope of the paper. 

\begin{figure}[t]
\centering
\bt{cc}
\hspace{-0.3cm}\includegraphics[width=0.5\textwidth]{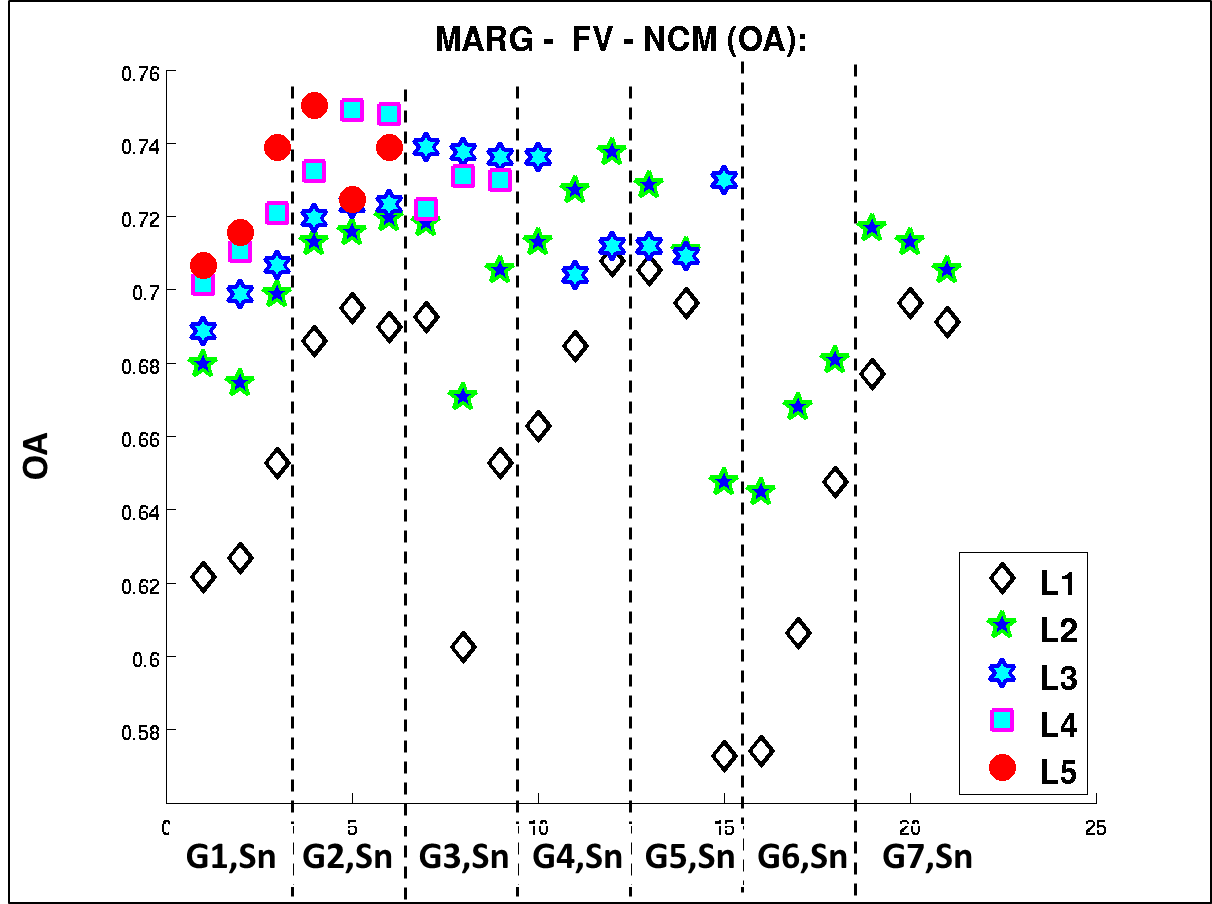} &
\hspace{-0.3cm}\includegraphics[width=0.5\textwidth]{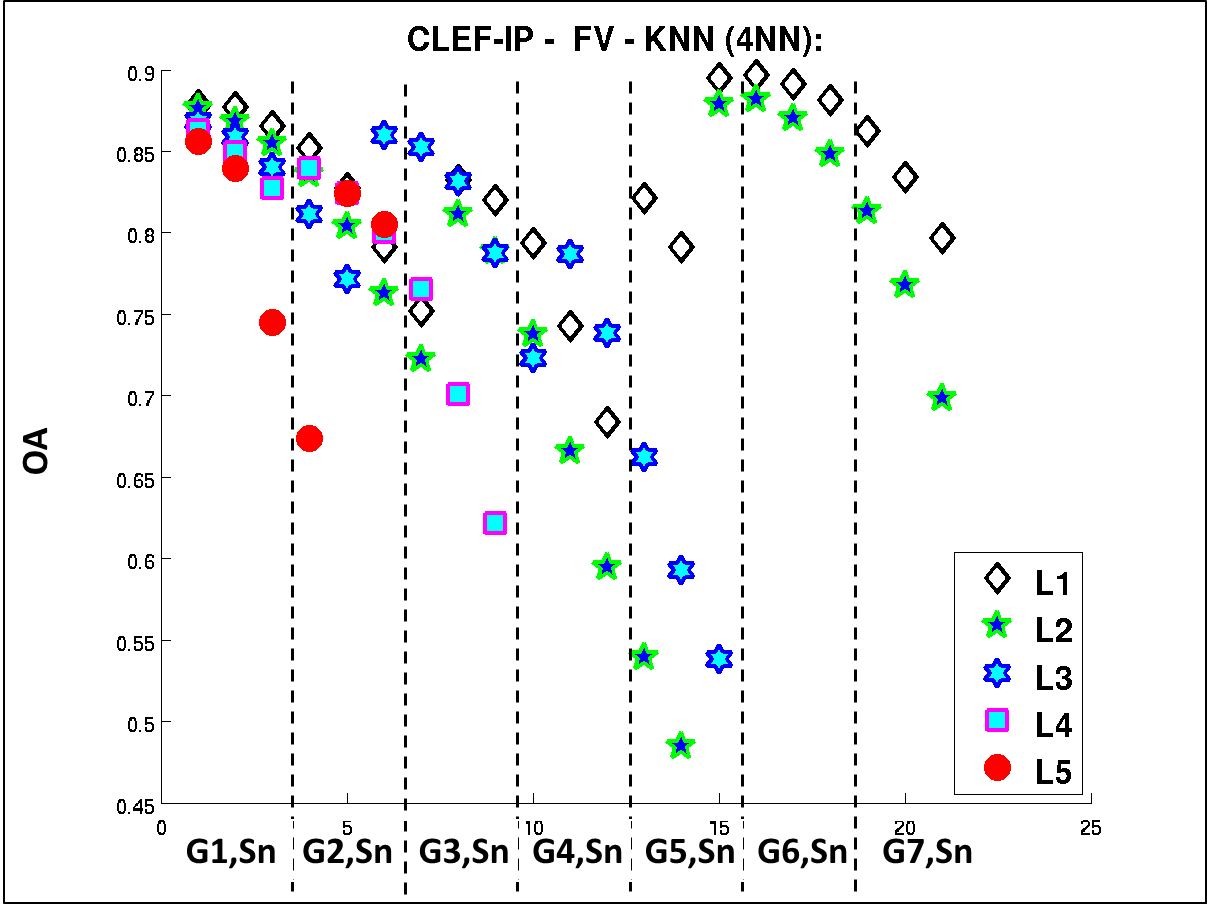}
\et 
\caption{Examples plots comparing different pyramid layers.}
\label{fig.Lcomp}
 \end{figure}

In Table \ref{tab:SLGcomp} and \fig{fig.Lcomp} we show  results 
when we vary the image size, the number of pyramid layers 
and the number of Gaussians and fix the other parameters  to
W3, F1 and M5. Analyzing the results suggest that while the best 
configuration varies a lot, in general we get best results with relatively 
few layers or even a single one and often only with few Gaussians. \\

{\bf Spatial pyramid.} If we analyze these results in more details,  
we can see that only NCM and SVM on  MARG 
performed best with 5 and 4 layers respectively. In general, it seems that NCM was the one that took 
the most advantage from more than 2 layers. In the case of SVM what is beneficial is 
large signatures (which is not surprising), but using fewer
layers with larger vocabularies seems to perform better than 
smaller vocabularies with more layers. This is somewhats in contrast 
to what we observed for RL and the results in \cite{krapac11iccv} concerning 
spatial pyramids with FVs on natural images.

In Table \ref{tab:SLGcomp} we also show results (in blue) for each dataset given a fixed configuration 
found as reasonably close to best results on most tasks. These configurations are (S3,L2,G4) for MARG and IH1,  
(S4,L2,G4) in the case of NIT and (S4,L1,G4) for CLEF-IP, where we have in addition (W3,F1,M5) for all datasets.
We can see that in most cases these fixed values are indeed good choses, except for  CLEF-IP for which it is
the less obvious  to find a good set of configurations, especially concerning the image size
that performs well  on all tasks (as shows also \fig{fig.MScomp}, bottom row).
This is probably  due to the fact that in this dataset the size of the images is extremely variable. 
The best  compromise we found was S4, G4 without spatial pyramid (L1), however the drop 
in accuracy is more important  than for  the other datasets.

\begin{table}[ttt]
\small
\begin{center}
\bt{|c||c|c|c|c||}
\hline
&  MARG & IH1   & NIT & CLEF-IP   \\
\hline
P@1 & \rk{95.1}/\bk{95.1}  & \rk{95.2}/\bk{94.5}  &  \rk{83.3}/\bk{81.7}  & \rk{89}/\bk{87.9}  \\
   & S3,L2,G4  & S3,L2,G1 & S4,L2,G4 & S4,L1,G1  \\
\hline
P@5 & \rk{83.1}/\bk{83.1} & \rk{93.8}/\bk{92.9}  & \rk{75.9}/\bk{74.9}  & \rk{86}/\bk{84.6} \\
  & S3,L2,G5 & S3,L2,G1  & S4,L3,G4  & S4,L1,G1 \\
\hline
MAP &  \rk{36.2}/\bk{36.2}   & \rk{76.5}/\bk{75.4} &  \rk{46.6}/\bk{41.6}  & \rk{46.1}/\bk{42.1}  \\
 &   S3,L2,G5  &  S3,L2,G1 &  S4,L4 ,G1 &  S2,L1,G1 \\
\hline
\hline
KNN & \rk{93.1}/\bk{93.1}   & \rk{95}/\bk{94.2}  &  \rk{82.8}/\bk{81.9}  & \rk{89.6}/\bk{88.1}  \\
 & S3,L2,G3 & S3,L2,G2   &   S4,L3,G4 &  S4,L1,G1   \\
\hline
NCM &  \rk{75}/\bk{72.4}  & \rk{92.4}/\bk{92.4}  &  \rk{78.3}/\bk{75.8}   & \rk{76.1}/\bk{75.1}   \\
 & S3,L5,G2 &  S3,L3,G1 & S4,L3,G5 & S2,L4,G1 \\
\hline
SVM & \rk{92.5}/\bk{90.7} & \rk{97.4}/\bk{97.2}  & \rk{86.5}/\bk{84.9}   & \rk{95.5}/\bk{93.5} \\
 & S3,L4,G3 & S4,L1,G7 &  S3,L1,G4  &  S4,L2,G6\\
 \hline
\et 
\caption{Comparative FV results on different datasets and tasks. 
We show best results (red) versus result using fixed parameter settings (blue) where we
used (S3,L2,G4) for MARG and IH1,  (S4,L2,G4) in the case of NIT and
(S4,L1,G4) for CLEF-IP.  We also show the parameter 
setting that provided the best results.}
\label{tab:SLGcomp}
\end{center}
\end{table}

 Finally, to show the influence of the visual model, 
 we rerun the last set of experiments, but instead of using
 the visual models built on the XRCE dataset, for each dataset we trained its own model
 with the SIFT features extracted from the training images of the tested dataset. 
 The results in Table~\ref{tab:SLGown} show on one hand that  we do not have a clear winner
 between the two models. On the other hand while the best configuration per dataset and task varies, 
 the  best score obtained are often close.  This show somewhat that the data on which the vocabulary is built has relatively
 little influence on the results the moment we use the images that have similar content which is the case for document images.

 \begin{table}[ttt]
\small
\begin{center}
\bt{|c||c|c|c|c||}
\hline
&  MARG & IH1   & NIT & CLEF-IP   \\
\hline
P@1 & \rk{95.9}/\bk{95.1}  & \rk{95.4}/\bk{95.2}  &  \rk{82.5}/\bk{83.3}  & \rk{87.7}/\bk{89}  \\
  & S3,L5,G1  & S4,L2,G1 & S4,L2,G3 & S4,L1,G1  \\
\hline
P@5 & \rk{85}/\bk{83.1} & \rk{94}/\bk{93.8} & \rk{75.4}/\bk{75.9}  & \rk{84.7}/\bk{86} \\
  & S4,L4,G2  & S4,L2,G1 & S4,L3,G2 & S4,L1,G1  \\
\hline
MAP &  \rk{37.8}/\bk{36.2}   & \rk{76.6}/\bk{75.4} &  \rk{45.8}/\bk{46.1}  & \rk{42.8}/\bk{46.1}  \\
 &   S3,L4,G2  &  S4,L2,G1 &  S4,L4,G1 &  S2,L1,G1 \\
\hline
\hline
KNN & \rk{95}/\bk{93.1}   & \rk{95.3}/\bk{95}  &  \rk{82.8}/\bk{82.8}  & \rk{88.8}/\bk{89.1}  \\
	& S3,L3,G2 & S4,L3,G1   &   S4,L3,G2 &  S4,L1,G1   \\
\hline
NCM &  \rk{81.1}/\bk{75}  & \rk{92.6}/\bk{92.4}  &  \rk{78.7}/\bk{78.3}   & \rk{75.7}/\bk{76.1}  \\
 & S2,L3,G4 &  S4,L3,G1 & S4,L4,G1 & S2,L4,G1 \\
\hline
SVM & \rk{94.1}/\bk{92.5} & \rk{97.4}/\bk{97.4}  & \rk{86.5}/\bk{86.5}   & \rk{95.8}/\bk{95.5} \\
 & S4,L5,G2 & S4,L2,G7 &  S3,L1,G4  &  S4,L2,G6\\
\hline
\et 
\caption{FV results when the visual models  were built 
SIFT features extracted from the images of the tested dataset (red)
and compared with the visual model on XRCE (blue). We also show the parameter 
setting that provided the best results for the results obtained with the visual models trained on 
the dataset itself.}
\label{tab:SLGown}
\end{center}
\end{table}

\subsection{Combine RL with  FV}
\label{sec:comb_rl_fv}

The most natural way to combine RL and FV is early or late fusion. As we use
dot product for retrieval, the dot product of the concatenated features (early fusion) is equivalent 
to the sum of the dot products (late fusion). Similarly, the NCM centroids of the 
concatenated features are the concatenation of the RL respectively FV centroids, and from 
therefore late and early fusion are again equivalent. 

To test the late fusion of RL with FV, we consider the configuration (S0,L5,Q11)
for RL and the fixed parameters settings leading to the values in blue  in the Table \ref{tab:SLGcomp} for FV. 
The results obtained are shown in Table \ref{tab:RL_FV}. We can see
 that even with a simple equally weighted late fusion,  in general (except for NIT) we 
obtain significant improvements both on retrieval and classification.

We would like to mention here another possible combinations of the RL with FV where
the main idea is to consider the RL features as 
low level features (replacing the SIFT) such that on each local window we build a RL histogram.
Then  the visual vocabulary (GMM) and the FV are built with these  local RL features directly or as
some PCA reduced forms of them.  Note also that if we use 
small image patches, the number of quantization of the runs (Q) can be reduced
 as anyway a run cannot be longer than the patch size. We intend in the future to explore 
 if such FVs on RL performs better than the global RL features and also if combining all 
 three signatures can further improve the accuracy.

 \section{Image-based Patent Retrieval}
 \label{sec:challenge}

\begin{table}[ttt]
\small
\begin{center}
\bt{|c|c||c|c|c|c||}
\hline
 & signature  &  MARG & IH1   & NIT & CLEF-IP   \\
\hline
MAP & RL & 33.2  &  64.3 &  38.5 &  35.6\\
	& FV &  36.1  & 75.4 &  41.6 & 42.1 \\
& RL+FV  & \rk{36.5} &  \rk{76.9} & \rk{44.2} & \rk{48.2} \\
\hline
KNN & RL  & 89.9  & 93.1 & 76.7 &  80.8\\
	 & FV & 92.6 & 94.2 & \rk{81.9} & 88.1  \\
	& RL+FV & \rk{93} & \rk{95.2} & 80.2 &  \rk{91.1} \\
\hline
NCM & RL  & 63.2  & 91.3 & 65.6  &  61.2 \\
& FV  &  72.4 & 92.4 &  75.8 & 75.1 \\
& RL+FV  &  \rk{73.4}  & \rk{93} &  \rk{79.8} &  \rk{75.8}\\
\hline
SVM & RL & 91.9 &  96.7 &   78.2 & 89.5 \\
 & FV & 90.7 & 97.2 & 	\rk{84.9}  & 93.5\\
 & RL+FV & \rk{92.8} & \rk{97.7} &  83.7  & \rk{94.4} \\
\hline
\et
\caption{Results with late fusion of RL and FV features on different datasets and tasks.}
\label{tab:RL_FV}
\end{center}
\end{table}

We would like first to recall briefly our participation in the Image-based Patent Retrieval task's
at Clef-IP 2011~\cite{Piroi11}. A more detailed description especially concerning the text 
representation and retrieval can be found in  \cite{clefip11}. The aim  of the challenge 
was to rank patents as relevant or non relevant one given a query patent while  using both visual 
and textual information.  There were 211 query patents provided and the collection to search in
contained 23444 patents  having  an application date previous to 2002. 
The number of images varied a lot, from few images to several hundred of images
per patent. In  total we had 4004 images  
in the query patents  and 291,566 images in the collection. 
 As image representation  we used the FV with the configuration  (S3, W2, F2, G5 and L1) where the model (PCA and GMM) were
  trained on CLEF-IP, \ie the training set of the  Image Classification Task of 
Clef-IP 2011~\cite{Piroi11}. The similarity between images was given by the dot product of two 
Fisher Vectors.  

We tested two main strategies.  In the first case, we 
considered the average distance between all pairs of images  given 
two patents with the corresponding set of images (MEAN). In the second case  
we considered only the maximum of all similarities computed between pairs of images (MAX). 

We also considered to integrate in the system our   automatic 
image-type classifier (using the same FV features) that was trained on the CLEF-IP dataset
and we used it to  predicted the image type. Using the predicted scores we
considered the similarities between class means 
(averaging the images predicted to belong to a given class), and took the 
average or the maximum according to the  strategy considered.

Finally, as in 
the considered patent classes (A43B patents related to  footwear, 
A61B patents concerning diagnoses and surgery and 
H01L patents proposing new semiconductor and electric solid state devices)
the drawings were the most relevant images, we discarded all images not predicted as
drawings and computed the mean or max similarities between the images predicted as drawings.  
Note that for other patent classes, considering images containing 
chemical structures or gene sequences would be more appropriate.

The results detailed in~\cite{clefip11} are recalled in Table~\ref{tab:visruns}. They show that
the max strategy is better than considering average similarities. 
Considering class means instead of global mean improves the MEAN 
strategy, but has no effect on the max strategy. Finally, considering only drawings performed the best
for both strategies.

While all these  retrieval accuracies are very low, 
we want to make a few remarks. First, the task was really challenging as relevant prior art 
patents do not necessary contain  images similar to relevant images in the query patent. 
Second, even with this poor  image based ranking 
and simple late fusion we were  able to improve the  text only based patent ranking especially with 
the I5 strategy (see details in~\cite{clefip11}).  Third, we can use more complex fusion methods 
to merge visual and textual  retrieval (see \eg the 
graph based  methods  described  in~\cite{AhpineTOIS15}).

\begin{table}[ttt]
	\caption{Image-based Patent Retrieval: overview of the performances of 
our different approaches. The performances are all shown in percentages.}	
	\label{tab:visruns}
	\begin{center}
\begin{tabular}{|c||c||c|}
\hline
 Model /strategy  & MEAN & MAX \\
\hline
\begin{tabular}{l}
Classifier \\
\hline
not used \\
class means  \\
only drawings 
\end{tabular} & 
\begin{tabular}{c|c|c}
ID & MAP  & P@10   \\
\hline
I1 & 0.56 & 0.20 \\
I3 & 0.80 & 0.40 \\
I5 & 1.09 & 0.62
\end{tabular} & 
\begin{tabular}{c|c|c}
ID & MAP  & P@10 \\
\hline
I2 & 1.84 & 0.75 \\
I4 & 1.84 & 0.70 \\
I6 & {\bf 3.51} & {\bf 1.85}
\end{tabular}\\
\hline
\end{tabular}
\end{center}
\end{table}

On the other hand the image type classification can also be improved in several way. 
On one hand we can  select  better feature configuration for FV combined with 
RL features as above or even using some new, deep convolutional neural networks (CNNs) based representations 
such as in  \cite{lekang14,harleyetal15}. 

Second,  the strategies to consider and combine features from 
set of images in~\cite{clefip11} was rather simple.  Instead, we can see
the set of patent images as a multi-page document and use the 
methods proposed in \cite{GordoPerronninICPR10,Gordo13,Rusnol14} to handle 
classification and retrieval with multi-page documents. 

For example, the bag-of-pages model of \cite{GordoPerronninICPR10}, consider PCA-reduced 
RL features  for  each page and build a FV for the document, \ie  
when computing FV  with \eq{eqn:ds}, the features 
 $\bm x_t$ corresponds to the RL features computed for the pages in the document page.
Similarly, we can build a FV with the RL features built on the patent images, 
and represent the patent containing these images with the obtained FV. 
 Then two patents are compared with the dot product of these  FVs. 

In \cite{Gordo13} the bag-of-classemes was proposed and have been shown to outperform the
bag-of-pages. In this case the $\bm x_t$ features are 
the image type classification scores concatenated into a single vector (called classeme)
and the FV is built on top of these vectors. Note nevertheless, that 
while in~\cite{Gordo13} the bag-of-classemes outperforms the bag-of-pages, the addressed problem is
different, \ie document classification. In addition all pages have the same class label, 
the one inherited 
from the document. In our case, in a patent we have different image types and therefore we could describe by
bag-of-classemes  the distribution  of different type of  images within a patent. 
While this can be a  useful information for the patent expert, it does not necessarily improves 
for example patent prior art search.  

Finally, we can also improve the image type classification by combining the 
visual information with information from text. Text can come from the patent, if we can
 access  image caption and/or the paragraphs  where  the image is referred.  
 The extracted text can be represented by bag-of-words that can be used to train  classifiers
which learn implicitly which  words are relevant to discriminate  image types. The textual 
and visual classifiers can after be merged at score level (late fusion). Alternatively, we can consider 
embedding both the visual and textual features in the same subspace using CCA 
and train a classifier in the embedded space as in~\cite{GordoDAS12}.
 Note that text information can also be extracted from the document
 image using OCR.  In the case of patent images, using  bag of "n-grams of characters" 
 on the text extracted from the image content could be more appropriate than bag-of-words 
to describe for example gene sequences,  mathematical formulas and  chemical structures.


\section{Conclusion}
\label{sec:conclusion}

In this paper we made an exhaustive experimental study on
RunLenght Histogram (RL) and Fisher Vector (FV) based  representations  
for document image classification  and retrieval. 
We compared different parameter configurations for both features using 
several datasets, methods and evaluation methods. We designed suitable configurations 
for both features and  while they might be suboptimal for individual 
tasks, features designed with the proposed configurations are reasonable in case 
one might want to solve different tasks with the same features. Then we discussed 
the usage of patent images in  prior art search as such an example.

\bibliographystyle{plain}
\bibliography{docimgs}

\end{document}